\documentclass[twoside,11pt]{article}
\usepackage[preprint]{jmlr2e}
\usepackage[utf8]{inputenc}
\usepackage[T1]{fontenc}
\usepackage[english]{babel}
\usepackage{verbatim}
\usepackage{url}
\usepackage{lmodern}
\usepackage{microtype}

\usepackage{xspace}
\usepackage{enumitem}
\setitemize{itemsep=0pt}
\usepackage[toc,page,titletoc]{appendix}
\usepackage[nottoc]{tocbibind}
\settocbibname{References}
\usepackage{lastpage}
\usepackage{listings}
\usepackage{silence}
\usepackage[font={small}]{caption}
\usepackage{subcaption}
\usepackage{graphicx}
\usepackage{tikz}
\usepackage{wrapfig}
\usepackage{amssymb}
\usepackage{pifont}
\usetikzlibrary{shapes}
\usepackage{algorithm}
\usepackage{algpseudocode}

\usepackage{amsfonts}
\usepackage{amsmath}
\usepackage{xfrac}
\usepackage{bm}
\usepackage{bbold}
\usepackage{mathtools}
\usepackage{wasysym}

\usepackage{booktabs}
\usepackage{soul}
\PassOptionsToPackage{table}{xcolor}
\usepackage{changepage,threeparttable}
\usepackage{tabularx}
\usepackage{multirow}
\usepackage{etoolbox}
\usepackage{makecell}
\usepackage{xtab}
\usepackage{tabularx}
\usepackage{placeins}
\usepackage{colortbl}
\definecolor{dkgreen}{rgb}{0,0.6,0}
\definecolor{dkred}{rgb}{0.8,0.0,0}
\newcommand{\cmark}{\color{dkgreen}\ding{51}}
\newcommand{\xmark}{\color{dkred}\ding{55}}
\usepackage{lscape}
\usepackage{hyperref}
\definecolor{customblue}{rgb}{0,0.08,0.45}
\definecolor{customgreen}{RGB}{85,107,47}
\definecolor{custompink}{RGB}{170,51,106}
\definecolor{egyptianblue}{rgb}{0.06, 0.2, 0.65}
\usepackage[nameinlink]{cleveref}
\creflabelformat{equation}{#2#1#3}
\hypersetup{
    colorlinks=true,
    linkcolor=customblue,
    filecolor=magenta,
    urlcolor=custompink,
    citecolor=egyptianblue
} 

\DeclareMathOperator*{\argmin}{arg\,min}					
\newcommand{\E}{\mathbb{E}} 		                        

\newcommand{\R}{\mathbb{R}}					

\DeclareMathAlphabet{\mathsfit}{\encodingdefault}{\sfdefault}{m}{sl}
\SetMathAlphabet{\mathsfit}{bold}{\encodingdefault}{\sfdefault}{bx}{n}

\newcommand{\eg}{\textit{e}.\textit{g}.}
\newcommand{\ie}{\textit{i}.\textit{e}.}
\newcommand{\methods}{\textsc{Source}\xspace}

\newcommand{\method}{\textsc{Source}}

\newcommand{\loo}{\textsc{LOO}}
\newcommand{\loos}{\textsc{LOO}\xspace}

\newcommand{\trainingData}{\mathcal{D}}
\newcommand{\trainingDataSubset}{\mathcal{S}}
\newcommand{\dataPoint}{\boldsymbol{z}}
\newcommand{\params}{\boldsymbol{\theta}}
\newcommand{\optParams}{\boldsymbol{\theta}^\star}
\newcommand{\paramsCheckpoint}{\hat{\boldsymbol{\theta}}}
\newcommand{\finalParams}{\boldsymbol{\theta}^s}
\newcommand{\hyperParams}{\boldsymbol{\lambda}}
\newcommand{\randomness}{\xi}

\newcommand{\hyper}{\boldsymbol{\lambda}}
\newcommand{\loss}{\mathcal{L}}
\newcommand{\cost}{\mathcal{J}}
\newcommand{\measurement}{f}
\newcommand{\attrib}{\tau}
\newcommand{\response}{r}

\newcommand{\eye}{\mathbf{I}}
\newcommand{\LR}{\eta}

\newcommand{\weight}{\epsilon}
\newcommand{\hessian}{\mathbf{H}}

\newcommand{\jacobian}{\mathbf{J}}

\newcommand{\grad}{\mathbf{g}}

\newcommand{\batch}{\mathcal{B}}
\newcommand{\segment}{\mathbf{S}}

\newcommand{\segmentS}{\mathbf{S}}
\newcommand{\segmentR}{\mathbf{r}}
\newcommand{\barSegmentS}{\bar{\segmentS}}
\newcommand{\barHess}{\bar{\hessian}}
\newcommand{\barGrad}{\bar{\grad}}
\newcommand{\barLR}{\bar{\LR}}

\newcommand{\numCheckpoint}{C}
\newcommand{\numUpdate}{T}
\newcommand{\numSegment}{L}
\newcommand{\numBatch}{B}
\newcommand{\numParam}{D}
\newcommand{\numData}{N}
\newcommand{\precond}{\mathbf{P}}
\newcommand{\precondApprox}{\bar{\precond}}

\begin{document}

\title{Training Data Attribution via \\ Approximate Unrolled Differentiation}

\author{%
    \name Juhan Bae$^{1,2}$ \email jbae@cs.toronto.edu \\
    \name Wu Lin$^2$ \email wu.lin@vectorinstitute.ai \\
    \name Jonathan Lorraine$^{1,2,3}$  \email lorraine@cs.toronto.edu \\
    \name Roger Grosse$^{1,2,4}$  \email rgrosse@cs.toronto.edu\AND
    \addr $^1$University of Toronto;
    \addr $^2$Vector Institute;
    \addr $^3$NVIDIA;
    \addr $^4$Anthropic
}

\editor{null}

\maketitle

\begin{abstract}%
Many training data attribution (TDA) methods aim to estimate how a model's behavior would change if one or more data points were removed from the training set. Methods based on implicit differentiation, such as influence functions, can be made computationally efficient, but fail to account for underspecification, the implicit bias of the optimization algorithm, or multi-stage training pipelines. By contrast, methods based on unrolling address these issues but face scalability challenges. In this work, we connect the implicit-differentiation-based and unrolling-based approaches and combine their benefits by introducing \textsc{Source}, an approximate unrolling-based TDA method that is computed using an influence-function-like formula. While being computationally efficient compared to unrolling-based approaches, \textsc{Source} is suitable in cases where implicit-differentiation-based approaches struggle, such as in non-converged models and multi-stage training pipelines. Empirically, \textsc{Source} outperforms existing TDA techniques in counterfactual prediction, especially in settings where implicit-differentiation-based approaches fall short.
\end{abstract}

\jmlrheading{23}{2023}{1-\pageref{LastPage}}{1/21; Revised 5/22}{9/22}{21-n0000}{Juhan Bae, Wu Lin, Jonathan Lorraine, Roger Grosse}
\ShortHeadings{Training Data Attribution via Approximate Unrolled Differentiation}{J. Bae, W. Lin, J. Lorraine \& R. Grosse}

\section{Introduction}
\label{sec:introduction}

Training data attribution (TDA) techniques are motivated by understanding the relationship between training data and the properties of trained models. TDA methods identify data points that significantly influence a model's predictions, making them invaluable for interpreting, debugging, and improving models \citep{koh2017understanding,yeh2018representer,feldman2020neural,han2020explaining,ilyas2022datamodels,park2023trak,grosse2023studying,konz2023attributing}. These techniques also have diverse applications in machine learning, such as detecting mislabeled data points \citep{pruthi2020estimating,kong2021resolving,jiang2023opendataval}, crafting data poisoning attacks \citep{fang2020influence,jagielski2021subpopulation,oh2022rank}, and curating datasets \citep{liu2021influence,xia2024less,engstrom2024dsdm}.

Many TDA methods aim to perform a \emph{counterfactual prediction}, which estimates how a trained model's behavior would change if certain data points were removed from (or added to) the training dataset. Unlike sampling-based approaches, which require repeated model retraining with different subsets of the dataset, gradient-based TDA techniques estimate an infinitesimal version of the counterfactual without model retraining. Two main strategies for gradient-based counterfactual TDA are \emph{implicit differentiation} and \emph{unrolled differentiation}.

Implicit-differentiation-based TDA, most notably influence functions \citep{hampel1974influence,koh2017understanding}, uses the Implicit Function Theorem \citep{krantz2002implicit} to estimate the optimal solution's sensitivity to downweighting a training data point. These methods are well-motivated for models with strongly convex objectives and provide convenient estimation algorithms that depend solely on the optimal model parameters rather than intermediate checkpoints throughout training. However, the classical formulation relies on assumptions such as uniqueness of and convergence to the optimal solution, which limits its applicability to modern neural networks \citep{basu2020influence,bae2022if,schioppa2023theoretical}.

By contrast, unrolling-based TDA, such as \textsc{SGD-Influence} \citep{hara2019data}, approximates the impact of downweighting a data point's gradient update on the final model parameters by backpropagating through the preceding optimization steps. Unrolling is conceptually appealing in modern neural networks because it does not rely on the uniqueness of or convergence to the optimal solution. Furthermore, it can incorporate details of the training process, such as the choice of optimizer, learning rate schedules, or a data point's position during training. For example, unrolling-based approaches can support TDA for multi-stage training procedures, such as in continual learning or foundation models, where the model undergoes multiple training phases with different objectives or datasets. However, they require storing all intermediate variables generated during the training process (\eg, parameter vectors for each optimization step) in memory for backpropagation, which can be prohibitively expensive for large-scale models. Notably, past works have considered applying unrolling to only the last epoch for large-scale models \citep{hara2019data,chen2021hydra}, restricting applicability in analyzing the effect of removing a data point at the beginning of training or in analyzing multi-stage training processes.

\begin{table}[]
    \centering
    \resizebox{\textwidth}{!}{%
    \begin{tabular}{@{}ccccc@{}}
    \toprule
    TDA Strategy & \begin{tabular}[c]{@{}c@{}}Number of  \\ Checkpoints \end{tabular} & \begin{tabular}[c]{@{}c@{}}Allows \\ Non-Convergence\end{tabular} & \begin{tabular}[c]{@{}c@{}}Supports \\ Multi-Stage\end{tabular} & \begin{tabular}[c]{@{}c@{}}Incorporates\\ Optimizer\end{tabular} \\ \midrule
    Implicit Differentiation \citep{koh2017understanding} & $1$ & \xmark & \xmark & \xmark \\
    Unrolled Differentiation \citep{hara2019data} & $\numUpdate$ & \cmark & \cmark & \cmark \\ 
    \methods \textbf{(ours)} & $~~~~~~~~~\numCheckpoint~(\ll T)$ & \cmark & \cmark & \cmark \\ \bottomrule
    \end{tabular}%
    }
    \vspace{-0.19cm}
    \caption{\small Comparison of implicit-differentiation-based TDA, unrolling-based TDA, and \method. \methods introduces practical algorithms that offer the advantages of unrolling-based techniques, requiring only a few checkpoints instead of all intermediate checkpoints throughout training. In our experiments, we use $6$ checkpoints ($\numCheckpoint = 6$) for \method, which is significantly smaller than the total number of gradient updates $\numUpdate$ performed during training, as required for unrolling-based methods.}
    \label{tab:tablular_comparison}
\end{table}

In this work, we connect implicit-differentiation-based and unrolling-based approaches and introduce a novel algorithm that enjoys the advantages of both methods. We start from the unrolled differentiation perspective and, after introducing suitable approximations, arrive at an influence-function-like estimation algorithm. While our method approximately coincides with influence functions in the simple setting of a deterministic objective optimized to convergence, it applies to more general settings where unrolling is typically required. Specifically, our method divides the training trajectory into one or more segments and approximates the distributions of gradients and Hessians as stationary within each segment. These segments may represent explicit training stages, such as in continual learning or foundation models, or changes in the Hessian and gradients throughout training. Hence, we call our method \methods (\textbf{S}egmented stati\textbf{O}nary \textbf{U}n\textbf{R}olling for \textbf{C}ounterfactual \textbf{E}stimation).

\methods inherits several key advantages from unrolling. Firstly, it allows the attribution of data points at different stages of training, providing a more comprehensive framework for TDA. Secondly, \methods can incorporate algorithmic choices into the analysis, accounting for learning rate schedules and the implicit bias of optimizers such as SGD \citep{robbins1951stochastic} or Adam \citep{kingma2014adam}. Lastly, it maintains a close connection with the counterfactuals, even in cases where the assumptions made in implicit-differentiation-based methods, such as the optimality of the final parameters, are not met. However, unlike unrolling, \methods does not require storing all intermediate variables generated during training; instead, it leverages only a handful of model checkpoints. The comparisons of \methods with implicit-differentiation-based and unrolling-based TDA methods are summarized in \Cref{tab:tablular_comparison}.

We evaluate \methods for counterfactual prediction across various tasks, including regression, image classification, text classification, and language modeling. Our method outperforms existing TDA techniques in approximating the effect of retraining the network without groups of data points and identifying training data points that would flip predictions on some test examples when trained without them. \methods demonstrates distinct advantages in scenarios where traditional implicit-differentiation-based methods fall short, such as models that have not fully converged or those trained in multiple stages. Our empirical evidence suggests that \methods is a valuable TDA tool in various scenarios.

\section{Background}
\label{sec:background}

Consider a finite training dataset $\trainingData \coloneq \{\dataPoint_i\}_{i=1}^{\numData}$. We assume that the model parameters $\params \in \R^\numParam$ are optimized with a gradient-based iterative optimizer, such as SGD, to minimize the empirical risk on this dataset:
\begin{align}
    \cost(\params, \trainingData) \coloneq \frac{1}{\numData} \sum_{i=1}^\numData \loss (\dataPoint_i, \params),
    \label{eq:erm}
\end{align}
where $\loss$ is the (twice-differentiable) loss function. We use the notation $\optParams (\trainingDataSubset)$ to denote the optimal solution obtained when the model is trained on a specific subset of the dataset $\trainingDataSubset \subseteq \trainingData$, and $\optParams \coloneq \optParams(\trainingData)$ to denote the optimal solution on the full dataset $\trainingData$.

In practice, it is common to employ parameters $\finalParams$ that approximately minimize the empirical risk (\eg, the result of running an optimization algorithm for $\numUpdate$ iterations), as obtaining the exact optimal solution for neural networks can be challenging and may lead to overfitting \citep{bengio2012practical}. When necessary, we use the notation $\finalParams (\trainingDataSubset; \hyperParams, \randomness)$ to indicate the final parameters obtained by training with the dataset $\trainingDataSubset$, along with hyperparameters $\hyperParams$ (\eg, learning rate and number of epochs) and random choices $\randomness$ (\eg, parameter initialization and mini-batch order). This notation explicitly acknowledges the dependence of the final parameters on various factors beyond the training dataset itself.

\subsection{Training Data Attribution}
\label{subsec:tda}

TDA aims to explain model behavior on a query data point $\dataPoint_q$ (\eg, test example) by referencing data points used to fit the model. The model behavior is typically quantified using a measurement $\measurement(\dataPoint_q, \params)$, selected based on metrics relevant to the analysis, such as loss, margin, or log probability. Given hyperparameters $\hyperParams$ and a training data point $\dataPoint_m \in \trainingData$, an attribution method $\attrib(\dataPoint_q, \dataPoint_m, \trainingData; \hyperParams)$ assigns a score to a training data point, indicating its \emph{importance} in influencing the expected measurable quantity $\E_{\randomness} \left[ \measurement(\dataPoint_q, \finalParams(\trainingData; \hyperParams, \randomness)) \right]$, where the expectation is taken over the randomness in the training process. In cases where an optimal solution to \Cref{eq:erm} exists, is unique, and can be precisely computed, and TDA is performed on this optimal solution, the attribution method is simply written as $\attrib(\dataPoint_q, \dataPoint_m, \trainingData)$.

One idealized TDA method is \emph{leave-one-out} (\loo) retraining \citep{weisberg1982residuals}, which assesses a data point's importance through counterfactual analysis. Assuming the above optimality condition is satisfied, for a chosen query data point $\dataPoint_q$ and a training data point $\dataPoint_m \in \trainingData$, the \loos score can be formulated as follows:
\begin{align}
    \attrib_{\loo} (\dataPoint_q, \dataPoint_m, \trainingData) \coloneq \measurement(\dataPoint_q, \optParams (\trainingData \setminus \{\dataPoint_m\})) - \measurement(\dataPoint_q, \optParams).\label{eq:loo}
\end{align}
When the measurement is defined as the loss, a higher absolute \loos score signifies a more substantial change in the query loss when the data point $\dataPoint_m$ is excluded from the training dataset, particularly when the model parameters are optimized for convergence. However, \loos retraining is computationally expensive, as it requires retraining the model for each training data point, making it infeasible for large models and datasets.

\subsection{Influence Functions}
\label{subsec:influence_functions}

Influence functions estimate the change in optimal parameters resulting from an infinitesimal perturbation in the weight of a training example $\dataPoint_m \in \trainingData$. Assuming that an optimal solution to \Cref{eq:erm} exists and is unique for various values of the data point's weight $\weight \in [-1, 1]$, the relationship between this weight and the optimal parameters is captured through the \emph{response function}:
\begin{align}
    \response(\weight) \coloneq \argmin_{\params} \cost(\params, \trainingData) + \frac{\weight}{\numData} \loss(\dataPoint_m, \params).
\end{align}
Influence functions approximate the response function using the first-order Taylor expansion around $\weight = 0$:
\begin{equation}
\begin{aligned}
    \response(\weight) \approx \response(0) + \frac{\mathrm{d} \response}{\mathrm{d} \weight} \Big|_{\weight=0} \cdot \weight = \optParams - \frac{\weight}{N} \hessian^{-1} \nabla_{\params} \loss(\dataPoint_m, \params^\star),
    \label{eq:if-param}
\end{aligned}
\end{equation}
where $\hessian \coloneq \nabla^2_{\params} \cost (\optParams, \trainingData)$ represents the Hessian of the cost function at the optimal solution, and the Jacobian of the response function $\sfrac{\mathrm{d} \response}{\mathrm{d} \weight} |_{\weight=0}$ is obtained using the Implicit Function Theorem \citep{krantz2002implicit}. The change in the optimal parameters due to the removal of $\dataPoint_m$ can be approximated by setting $\weight = -1$:
\begin{align}
    \optParams(\trainingData \setminus \{\dataPoint_m\}) - \optParams \approx \frac{1}{\numData} \hessian^{-1} \nabla_{\params} \loss(\dataPoint_m, \params^\star).
\end{align}
By applying the chain rule of derivatives, influence functions estimate the change in a measurable quantity for a query example $\dataPoint_q$ due to the removal of a training point $\dataPoint_m$ as:
\begin{align}
    \attrib_{\textsc{IF}} (\dataPoint_q, \dataPoint_m, \trainingData) \coloneq \nabla_{\params} \measurement(\dataPoint_q, \optParams)^\top \hessian^{-1} \nabla_{\params} \loss(\dataPoint_m, \optParams). \label{eq:if}
\end{align}

We refer readers to \citet{koh2017understanding} for detailed derivations and discussions of influence functions. As observed in \Cref{eq:if}, influence functions provide algorithms that only depend on the optimal parameters $\optParams$ (rather than intermediate checkpoints). However, when applied to neural networks, the connection to the counterfactual prediction is tenuous due to the unrealistic assumptions that the optimal solution exists, is unique, and can be found \citep{basu2020influence,bae2022if,schioppa2023theoretical}. In practice, the gradients and Hessian in \Cref{eq:if} are computed using the final parameters $\finalParams$ from a single training run instead of the optimal solution.

Moreover, influence functions cannot incorporate the details of the training procedure, such as the implicit bias of the optimizer and the point at which a training example $\dataPoint_m$ appeared during training \citep{guu2023simfluence,nickl2024memory}. They are, hence, unsuitable for analyzing the effect of removing a data point at various stages of training or performing TDA on multi-stage training procedures. For instance, consider a case where the model was sequentially trained with two datasets $\trainingData_1$ and $\trainingData_2$, such as in continual learning and foundation models, and one would like to investigate the impact of removing a data point $\dataPoint_m \in \trainingData_1$ that appeared in the first stage of training. Influence functions do not provide any mechanism to separate multiple stages of training, and when computed using the combined dataset $\trainingData = \trainingData_1 \cup \trainingData_2$, they inherently assume that the final parameters are optimal on both datasets. However, this assumption may not hold as the final model parameters may no longer be precisely optimal on the data points that appeared in the first stage due to catastrophic forgetting \citep{goodfellow2013empirical}.

\subsection{Evaluation of TDA Techniques}
\label{subsec:tda_evaluate}

Given the focus on counterfactual prediction in many TDA methods, \loos estimates, defined in \Cref{eq:loo}, are often considered a ground truth for evaluating these techniques. However, the computation of \loos scores in neural networks encounters several computational and conceptual challenges, as detailed in \Cref{app:loo}. For a robust and standardized measure for evaluating TDA techniques, we instead use the linear datamodeling score (LDS) from \citet{park2023trak} as well as subset removal counterfactual evaluation \citep{hooker2019benchmark,yeh2022first,ilyas2022datamodels,zheng2023intriguing,park2023trak,brophy2023adapting,singla2023simple,georgiev2023journey}.

\vspace{-0.1cm}
\paragraph{Linear Datamodeling Score (LDS).} A TDA method $\attrib$, as detailed in \Cref{subsec:tda}, assigns a score to each pair of a query and training data point. The inherently \emph{additive} nature of most TDA techniques allows for the computation of a group attribution score for a specific training data subset $\trainingDataSubset \subset \trainingData$. The importance of $\trainingDataSubset$ on the measurable quantity $\measurement$ is estimated by summing the individual scores attributed to each data point within this subset. The group attribution is expressed as follows:
\begin{align}
    g_\attrib (\dataPoint_q, \trainingDataSubset, \trainingData; \hyperParams) \coloneq \sum_{\dataPoint \in \trainingDataSubset} \attrib (\dataPoint_q, \dataPoint, \trainingData; \hyperParams).
    \label{eq:group_pred}
\end{align}

Consider $M$ random subsets $\{ \trainingDataSubset_j \}_{j=1}^M$ from the training dataset, each containing $\lceil \alpha N \rceil$ data points for some $\alpha \in (0, 1)$. Given a hyperparameter configuration $\hyperParams$ to train the model, the LDS for a query point $\dataPoint_q$ is defined as:
\begin{align}
    \text{LDS}_{\alpha}(\dataPoint_q, \attrib) \coloneqq \boldsymbol{\rho} \left( \{ \E_{\randomness} \left[ f(\dataPoint_q, \finalParams(\trainingDataSubset_j; \hyperParams, \randomness)) \right] : j \in [M] \}, \{g_\attrib (\dataPoint_q, \trainingDataSubset_j, \trainingData; \hyperParams) : j \in [M] \} \right),
    \label{eq:LDS}
\end{align}
where $\boldsymbol{\rho}$ represents the Spearman correlation \citep{spearman1987proof}. This expected measurable quantity is approximated by retraining the network $R$ times under different random choices. The final LDS is obtained by averaging the scores across many (typically up to $2000$) query data points. In our experiments, we use $100$ data subsets ($M = 100$) and conduct a maximum of $100$ retraining iterations ($R \in \{5, 10, 20, 100\}$) for each subset to compute the LDS. 

\vspace{-0.1cm}
\paragraph{Subset Removal Counterfactual Evaluation.} Subset removal counterfactual evaluation examines the change in model behavior before and after removing data points that are highly ranked by an attribution technique. For classification tasks, we consider $100$ test data points that are correctly classified when trained with the full dataset and, for each test data point, examine if removing and retraining without the top-$k$ \emph{positively} influential data points can cause misclassification on average (trained under different random choices).\footnote{The literature also uses terms such as \emph{helpful} \citep{koh2017understanding}, \emph{proponent} \citep{pruthi2020estimating}, and \emph{excitatory} \citep{yeh2018representer} to describe positively influential training data points.} By assessing the impact of removing influential data points on the model's performance, counterfactual evaluation provides a direct measure of the effectiveness of TDA techniques in identifying data points that significantly contribute to the model's behavior. 

\vspace{-0.1cm}
\paragraph{Downstream Task Evaluation.} TDA techniques have also been evaluated on their performance on downstream tasks, such as mislabeled data detection \citep{khanna2019interpreting,pruthi2020estimating,kim2023gex}, class detection \citep{hanawa2020evaluation,kwon2023datainf}, finding hallucinations in the training dataset \citep{ladhak2022contrastive}, and retrieving factual knowledge from the training dataset \citep{akyurek2022tracing}. These tasks can offer additional insights into the effectiveness and applicability of data attribution methods in practical scenarios. However, the connections between these tasks and counterfactual prediction are often unclear \citep{sogaard2021revisiting,park2023trak}, and it is uncertain whether algorithmic improvements in counterfactual prediction will directly result in improved performance on these downstream tasks.

\section{Methods}
\label{sec:methods}

In this section, we introduce \methods (\textbf{S}egmented stati\textbf{O}nary \textbf{U}n\textbf{R}olling for \textbf{C}ounterfactual \textbf{E}stimation), a gradient-based TDA technique that combines the advantages of implicit differentiation and unrolled differentiation. We motivate our approach from the unrolling perspective and, after introducing suitable approximations, arrive at an influence-function-like estimation algorithm. Finally, we describe a practical instantiation of \methods by approximating the Hessian with the Eigenvalue-corrected Kronecker-Factored Approximate Curvature (EK-FAC) \citep{george2018fast} parameterization.

\subsection{Motivation: Unrolling for Training Data Attribution}
\label{subsec:unroll}

Consider optimizing the model parameters using SGD with a fixed batch size $\numBatch$, starting from the initial parameters $\params_0$.\footnote{For an extension to preconditioned gradient updates, see \Cref{app:general}.} The update rule at each iteration is expressed as follows:
\begin{align}
\params_{k+1} \leftarrow \params_k - \frac{\LR_k}{\numBatch} \sum_{i=1}^{\numBatch} \nabla_{\params} \loss(\dataPoint_{ki}, \params_k),
\label{eq:sgd}
\end{align}
where $\LR_k$ denotes the learning rate for iteration $k$, $\batch_k$ denotes a mini-batch of examples drawn randomly with replacement from the training dataset $\trainingData$, $\dataPoint_{ki}$ is the $i$-th data point in $\batch_k$, and $\numUpdate$ denotes the total number of iterations.

We aim to understand the effect of removing a training data point $\dataPoint_m \in \trainingData$ on the terminal model parameters $\params_T$. We parameterize the weight of $\dataPoint_m$ as $1 + \weight$ for $\weight \in [-1, 1]$, where $\weight = 0$ corresponds to the original training run and $\weight = -1$ represents the removal of a data point. This parameterization results in the following update rule:
\begin{equation}
\begin{aligned}
    \params_{k+1}(\weight) \leftarrow \params_k(\weight) - \frac{\LR_k}{\numBatch} \sum_{i=1}^{\numBatch} \left( 1 + \delta_{ki} \weight \right) \nabla_{\params} \loss(\dataPoint_{ki}, \params_k (\weight)), \label{eq:param_sgd}
\end{aligned}
\end{equation}
where $\delta_{ki} \coloneq \mathbb{1}[\dataPoint_{ki} = \dataPoint_m]$ is the indicator function for having selected $\dataPoint_m$. For brevity, the dependence of $\params$ on $\weight$ will usually be suppressed.

\begin{figure*}[!t]
    \centering
    \resizebox{\textwidth}{!}{%
    \begin{tabular}{p{\textwidth}}
        \textbf{Unrolled Differentiation}\\
        \midrule
        \centering
        \includegraphics[width=1.05\textwidth]{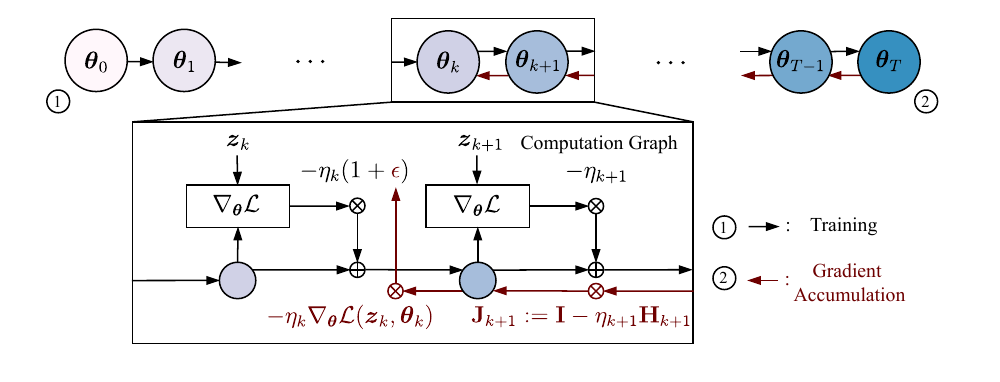}
    \end{tabular}
    }
    \vspace{-0.7cm}
    \caption{\small A simplified illustration of unrolled differentiation in SGD with a batch size of $1$ and a data point of interest $\dataPoint_m$ appearing once in training at iteration $k$. The highlighted nodes in the box represent the computation graph with the update rule from \Cref{eq:param_sgd}, where $B = 1$ and $\dataPoint_k = \dataPoint_m$. Unrolling backpropagates through the optimization steps from $\params_T$ to compute the total derivative with respect to ${\color[rgb]{0.4352,0,0} \epsilon}$, requiring all parameter vectors from $k$ to $T$ to be saved in memory.}
    \label{fig:computation_graph}
\end{figure*}

Similarly to other gradient-based TDA methods, such as influence functions, we approximate the change in the terminal parameters due to the data removal $\params_T(-1) - \params_T(0)$ with its first-order Taylor approximation $\sfrac{\mathrm{d} \params_T}{\mathrm{d} \epsilon} |_{\epsilon=0}$. Henceforth, we suppress the notation $|_{\epsilon=0}$ because this derivative will always be evaluated at $\epsilon=0$. The total derivative $\sfrac{\mathrm{d} \params_T}{\mathrm{d} \epsilon}$ can be evaluated by differentiating through the unrolled computation graph for the training procedure, as shown in \Cref{fig:computation_graph}. Let $\delta_k \coloneq \sum_{i=1}^{\numBatch} \delta_{ki}$ denote the number of times $\dataPoint_m$ is chosen in batch $\batch_k$. By applying the chain rule of derivatives, the contribution of iteration $k$ to the total derivative can be found by multiplying all the Jacobian matrices along the accumulation path (highlighted in {\color[rgb]{0.4352,0,0} red}), giving the value $-\tfrac{\LR_k}{B} \delta_k \jacobian_{k+1:T} \grad_k$, where:
\begin{equation}
  \begin{gathered}
    \jacobian_k \coloneq \frac{\mathrm{d} \params_{k+1}}{\mathrm{d} \params_k} = \eye - \LR_k \hessian_k \\
    \jacobian_{k:k'} \coloneq \frac{\mathrm{d} \params_{k'}}{\mathrm{d} \params_k} = \jacobian_{k'-1} \cdots \jacobian_{k+1} \jacobian_{k}\\
    \grad_k \coloneq \nabla_{\params} \loss(\dataPoint_m, \params_k).
  \end{gathered}
\end{equation}
Here, $\hessian_k \coloneqq \frac{1}{\numBatch} \sum_{i=1}^{\numBatch} \nabla^2_{\params} \loss(\dataPoint_{ki}, \params_k)$ is the mini-batch Hessian for iteration $k$ and we define $\jacobian_{k:k} \coloneqq \eye$ for any $0 \leq k < T$ by convention. 

\begin{wrapfigure}[19]{r}{0.4\linewidth}
    \centering
    \includegraphics[width=\linewidth]{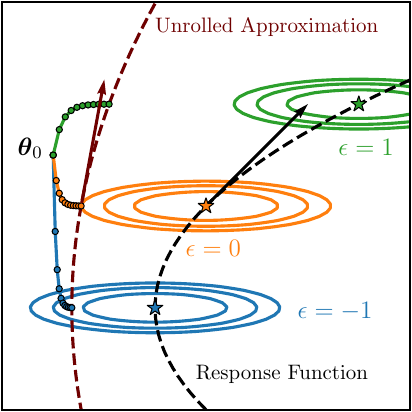}
    \vspace{-0.73cm}
    \caption{Illustrative comparision of influence functions and {\color[rgb]{0.4352,0,0} unrolling-based TDA}. Each contour represents the cost function at different values of $\epsilon$, which controls the degree of downweighting a data point $\dataPoint_m$.}
    \label{fig:unrolling}
\end{wrapfigure}
This unrolling-based formulation of TDA is advantageous in the context of modern neural networks. In contrast to influence functions (implicit-differentiation-based TDA; see \Cref{subsec:influence_functions}), unrolling does not assume uniqueness or convergence to the optimal solution. An illustrative comparison of the two approaches is shown in \Cref{fig:unrolling}. Exact influence functions differentiate the response function (\Cref{eq:if-param}), estimating the sensitivity of the optimal solution ($\star$) to downweighting a data point. By contrast, unrolling estimates the sensitivity of the \emph{final} model parameters (at the end of training) to downweighting a data point; hence, it can account for details of the training process such as learning rate schedules, implicit bias of optimizers, or a data point's position during training. For instance, in our illustrative example, gradient descent optimization is stopped early, such that the optimizer makes much progress in the high curvature direction and little in the low curvature direction. Unrolling-based TDA (but not implicit differentiation) accounts for this effect, resulting in a smaller influence along the low curvature direction.

The effect of removing $\dataPoint_m$ on any single training trajectory may be noisy and idiosyncratic. For stability, we instead consider the expectation over training trajectories, where the selection of training examples in each batch (and all downstream quantities such as the iterates $\params_k$) are treated as random variables.\footnote{We assume a fixed initialization $\params_0$ to break the symmetry.} We are interested in the average treatment effect $\E \left[ \params_T(-1) - \params_T(0) \right]$, where the expectation is over the batch selection, and approximate this quantity with $-\mathbb{E}\left[\sfrac{\mathrm{d} \params_T}{\mathrm{d} \epsilon}\right]$. The expected total derivative can be expanded as a sum over all iterations, applying linearity of expectation:
\begin{equation}
\begin{aligned}
    \E \left[ \frac{\mathrm{d} \params_{T}}{\mathrm{d} \epsilon} \right] 
    = \mathbb{E} \left[ -\sum_{k=0}^{T-1} \frac{\LR_k}{\numBatch} \delta_k \jacobian_{k+1:T} \grad_k \right] = -\sum_{k=0}^{T-1} \frac{\LR_k}{\numBatch} \E \left[ \delta_k \jacobian_{k+1:T} \grad_k \right].
    \label{eqn:unroll}
\end{aligned}
\end{equation}

In principle, we could compute a Monte Carlo estimate of this expectation by averaging many training trajectories. For each trajectory, $\sfrac{\mathrm{d} \params_T}{\mathrm{d} \epsilon}$ can be evaluated using reverse accumulation (\ie, backpropagation) on the computation graph. However, this approach is prohibitively expensive as it requires storing all intermediate optimization variables for the backward pass. Furthermore, many Monte Carlo samples may be required to achieve accurate estimates. 

\subsection{Segmenting the Training Trajectory}
\label{subsec:segment}

To derive a more efficient algorithm for approximating $\E \left[\sfrac{\mathrm{d} \params_T}{\mathrm{d} \epsilon} \right]$, we now partition the training procedure into $\numSegment$ segments and approximate the reverse accumulation computations for each segment with statistical summaries thereof. Our motivations for segmenting the training procedure are twofold. First, the training procedure may explicitly include multiple stages with distinct objectives and/or datasets, as in continual learning or foundation models. Second, the Hessians and gradients are likely to evolve significantly over training, and segmenting the training allows us to approximate their distributions as stationary within a segment (rather than over the entire training run).

We index the segments as $\ell = 1, \ldots, \numSegment$, with segment boundaries denoted as $T_\ell$. By convention, $T_L \coloneq T$ and $T_0 \coloneq 0$ denote the end of training and beginning of training, respectively, and $K_\ell \coloneq T_\ell - T_{\ell - 1}$ denotes the total number of iterations within a segment. Conceptually, we can compute the total derivative using reverse accumulation over a coarse-grained computation graph represented in terms of segments rather than individual iterations. The Jacobian associated with each segment is denoted as $\segmentS_\ell \coloneqq \jacobian_{T_{\ell-1}: T_{\ell}}$.

To approximate the expected total derivative $\E \left[ \sfrac{\mathrm{d} \params_T}{\mathrm{d} \weight} \right]$, we first rewrite \Cref{eqn:unroll} using the segment notation just introduced. We then approximate the Jacobians of different segments as statistically independent (see discussion below):
\begin{align}
    \E \left[ \frac{\mathrm{d} \params_{T}}{\mathrm{d} \weight} \right] 
    &= -\E \left[\sum_{\ell = 1}^L \sum_{k=T_{\ell-1}}^{T_{\ell} - 1} \frac{\LR_k}{B} \delta_k \left( \prod_{\ell'=L}^{\ell + 1}  \segmentS_{\ell'} \right) \jacobian_{k+1:T_{\ell}} \grad_k \right] \\
    &= -\E \Bigg[ \sum_{\ell=1}^L \left( \prod_{\ell'=L}^{\ell + 1} \segmentS_{\ell'} \right) \underbrace{\left( \sum_{k=T_{\ell - 1}}^{T_{\ell} - 1} \frac{\eta_k}{B} \delta_k \jacobian_{k+1:T_{\ell}} \grad_k \right)}_{\coloneq \segmentR_\ell} \Bigg]\\
    &\approx -\sum_{\ell=1}^L \left( \prod_{\ell'=L}^{\ell + 1} \E \left[ \segmentS_{\ell'} \right] \right) \E \left[\segmentR_\ell \right],
    \label{eqn:final}
\end{align}
where the last line uses our independence approximation to push the expectations inward. Note that our product notation $\prod_{\ell'=\numSegment}^{\ell + 1}$ takes $\ell'$ in decreasing order from $L$ down to $\ell + 1$.

To obtain tractable approximations for $\E[\segmentS_\ell]$ and $\E[\segmentR_\ell]$, we approximate the Hessian and gradients distributions as stationary within each segment. This implies that the Hessians within a segment share a common mean $\barHess_\ell \coloneq \E[\hessian_k]$ for $T_{\ell-1} \leq k < T_{\ell}$. Analogously, the gradients within a segment share a common mean $\barGrad_\ell \coloneq \E [\grad_k]$. Moreover, we approximate the step sizes within each segment with their mean $\barLR_\ell$. If these stationarity approximations are too inaccurate (\eg, $\E[\hessian_k]$ and/or $\mathbb{E}[\grad_k]$ change rapidly throughout the segment), one can improve the fidelity by carving the training trajectory into a larger number of segments, at the expense of increased computational and memory requirements. Finally, we approximate the Hessians and gradients in different time steps as statistically independent.\footnote{There are two sources of randomness in the gradient and Hessian at each step: the mini-batch sampling, and the optimization iterates (which, recall, we treat as random variables). Mini-batch sampling contributes to independent variability in different steps. However, autocorrelation of optimization iterates induces correlations between Hessians and gradients in different time steps. Our independence approximation amounts to neglecting these correlations.}

\vspace{-0.1cm}
\paragraph{Approximation of $\E [\segmentS_\ell]$.} We approximate $\E[\segmentS_\ell]$ in \Cref{eqn:final} as follows:
\begin{equation}
\begin{aligned}
    \E [\segmentS_\ell] = \E [\jacobian_{T_{\ell-1}:T_{\ell}}] \approx \left( \eye - \barLR_\ell \barHess_\ell \right)^{K_\ell} \approx \exp (-\bar{\LR}_\ell K_\ell \barHess_\ell) \coloneq \barSegmentS_\ell,
    \label{eq:S}
\end{aligned}
\end{equation}
where the first approximation uses the stationary and independence approximations and the second approximation uses the definition of matrix exponential.\footnote{Given a square matrix $\mathbf{M}$, the exponential of $\mathbf{M}$ is defined as $\exp(\mathbf{M}) = \lim_{k\to\infty} \left(\eye + \sfrac{\mathbf{M}}{k} \right)^k$.} One can gain an intuition for $\barSegmentS_\ell$ in \Cref{eq:S} by observing that it is a matrix function of $\barHess_\ell$.\footnote{Given a scalar function $F$ and a square matrix $\mathbf{M}$ diagonalizable as $\mathbf{M} = \mathbf{P} \mathbf{D} \mathbf{P}^{-1}$, the matrix function is defined as $F(\mathbf{M}) = \mathbf{P} F(\mathbf{D}) \mathbf{P}^{-1}$, where $F(\mathbf{D})$ applies $F$ to each diagonal entry of $\mathbf{D}$.} Let $\barHess_\ell = \mathbf{Q} \boldsymbol{\Lambda} \mathbf{Q}^\top$ be the eigendecomposition of $\barHess_\ell$ and let $\sigma_j$ be the $j$-th eigenvalue of $\barHess_\ell$. The expression in \Cref{eq:S} can be seen as applying the function $F_{\mathbf{S}}(\sigma) \coloneq \exp(-\bar{\LR}_\ell K_\ell \sigma)$ to each of the eigenvalues $\sigma$ of $\barHess_\ell$. The value is close to zero in high-curvature directions, so the training procedure ``forgets'' the components of $\params$ which lie in these directions. However, information about $\params$ is retained throughout the $\ell$-th segment for low-curvature directions.

\vspace{-0.1cm}
\paragraph{Approximation of $\mathbb{E}[\segmentR_\ell]$.} We further approximate $\mathbb{E}[\segmentR_\ell]$ in \Cref{eqn:final} as follows:
\begin{align}
    \mathbb{E}[\segmentR_\ell] &= \E \left[ \sum_{k=T_{\ell-1}}^{T_\ell - 1} \frac{\eta_k}{B} \delta_k \jacobian_{k+1:T_\ell} \mathbf{g}_k\right]\\
    &\approx \frac{1}{N} \sum_{k=T_{\ell-1}}^{T_{\ell} - 1} \bar{\LR}_\ell (\eye - \bar{\LR}_\ell \barHess_\ell)^{T_{\ell} - 1 - k} \barGrad_\ell \label{eqn:stationary_approx}\\
    &= \frac{1}{N}(\eye - (\eye - \bar{\LR}_\ell \barHess_\ell)^{K_{\ell}} )\barHess_\ell^{-1} \barGrad_\ell \label{eqn:finite}\\
    &\approx \frac{1}{N} \underbrace{(\eye - \exp(-\bar{\LR}_\ell K_\ell \barHess_\ell)) \barHess^{-1}_\ell}_{\coloneq F_{\mathbf{r}}(\sigma)} \barGrad_\ell \coloneq \bar{\mathbf{r}}_\ell, \label{eq:exp_approx}
\end{align}
where \Cref{eqn:stationary_approx} uses the stationary and independence approximations and $\mathbb{E}[\delta_k] = \sfrac{B}{N}$, \Cref{eqn:finite} uses the finite series,\footnote{For a symmetric square matrix $\mathbf{M}$, we have $\sum_{i=0}^{T-1} \mathbf{M}^i = (\eye - \mathbf{M}^T)(\eye - \mathbf{M})^{-1}$. When $\eye - \mathbf{M}$ is singular, we can replace $(\eye - \mathbf{M})^{-1}$ with the pseudoinverse $(\eye - \mathbf{M})^{+}$.} and, similarly to \Cref{eq:S},  \Cref{eq:exp_approx} uses the definition of the matrix exponential. We again observe that, because all the matrices commute, $\bar{\segmentR}_\ell$ in \Cref{eq:exp_approx} can be written in terms of a matrix function, defined as:
\begin{align}
    F_{\mathbf{r}} (\sigma) \coloneq \frac{1 - \exp{\left(-\bar{\LR}_\ell K_\ell \sigma\right)}}{\sigma}.
\end{align}
\begin{wrapfigure}[12]{r}{0.45\linewidth}
    \vspace{-0.34cm}
    \centering
    \includegraphics[width=\linewidth]{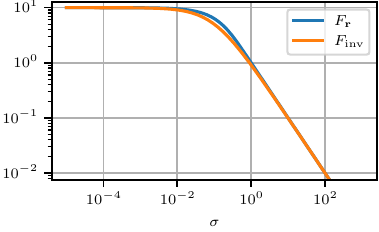}
    \vspace{-0.8cm}
    \caption{A demonstration of the match in qualitative behavior between $F_{\mathbf{r}}$ and $F_{\rm inv}$, where we set $\bar{\eta}_\ell = 0.1$ and $K_\ell = 100$.}
    \label{fig:matrix_function}
\end{wrapfigure}
In high-curvature directions, this term approaches $\sfrac{1}{\sigma}$, whereas in low-curvature directions, the formulation approaches to $\bar{\LR}_\ell K_\ell$. The qualitative behavior of $F_{\mathbf{r}}$ can be captured with the function $F_{\rm inv}(\sigma) \coloneq 1/(\sigma + \lambda)$, where $\lambda = \barLR^{-1}_\ell K_{\ell}^{-1}$, as shown in \Cref{fig:matrix_function}. Applying this to $\barHess_\ell$ results in approximating \Cref{eq:exp_approx} with the damped inverse Hessian-vector product $(\barHess_\ell + \lambda \eye)^{-1} \barGrad_\ell$. This is essentially the formula for influence functions, except that $\barHess_\ell$ and $\barGrad_\ell$ represent the expected Hessian and gradient rather than the terminal one, and our analysis yields an explicit formula for the damping parameter $\lambda$ (which would otherwise need to be hand-tuned). Hence, influence functions are approximately a special case with only a single segment, so our damped unrolling analysis gives an alternative motivation for influence functions. 

\begin{figure*}[!t]
    \vspace{-0.35cm}
    \centering
    \resizebox{\textwidth}{!}{%
    \begin{tabular}{p{\textwidth}}
        \textbf{\method}\\
        \midrule
        \includegraphics[width=1.05\textwidth]{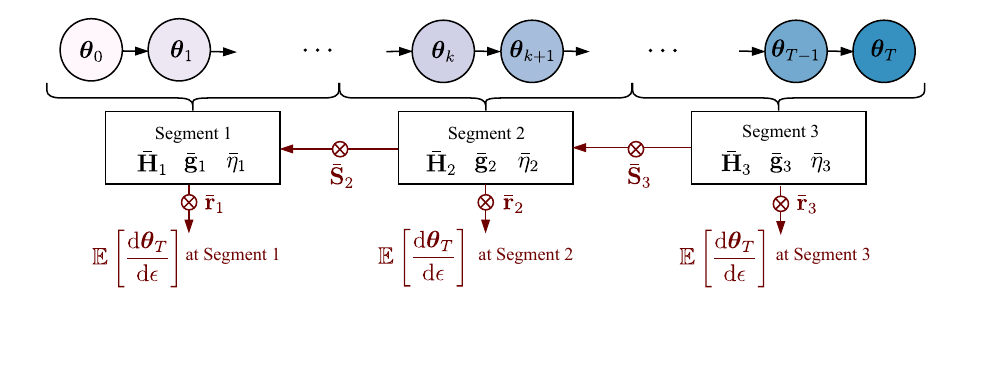}
    \end{tabular}
    }
    \vspace{-1.7cm}
    \caption{A simplified illustration of \methods with $3$ segments ($L = 3$), as defined in \Cref{eq:unif_segment}. \methods divides the training trajectory into one or more segments and approximates the gradient $\barGrad_\ell$ and Hessian $\barHess_\ell$ distributions as stationary with a fixed learning rate $\barLR_\ell$ within each segment $\ell$. Compared to unrolling in \Cref{fig:computation_graph}, \methods does not require storing the entire optimization variables throughout training. Instead, it only requires a handful of checkpoints throughout training to approximate the means of the Hessians and gradients.}
    \label{fig:source}
\end{figure*}

\subsection{Full Procedure}

Putting it all together, we derive a closed-form term to approximate the expected total derivative in \Cref{eqn:unroll}:
\begin{equation}
    \begin{aligned}
    \mathbb{E} \left[ \frac{\mathrm{d} \params_{T}}{\mathrm{d} \epsilon} \right] \approx -\frac{1}{N} \sum_{\ell=1}^L \left( \prod_{\ell'=L}^{\ell + 1} \bar{\segment}_{\ell'} \right) \bar{\mathbf{r}}_\ell,
    \label{eq:unif_segment}
\end{aligned}
\end{equation}
where $\bar{\mathbf{S}}_\ell$ and $\bar{\mathbf{r}}_\ell$ are obtained with \Cref{eq:S} and \Cref{eq:exp_approx}, respectively, and the expectation accounts for the average effect of downweighting a data point throughout training. We term our algorithm \methods (\textbf{S}egmented stati\textbf{O}nary \textbf{U}n\textbf{R}olling for \textbf{C}ounterfactual \textbf{E}stimation) and refer readers to \Cref{fig:source} for a visual illustration.

Similarly to unrolling, \methods can incorporate fine-grained information about optimization trajectories into the analysis. For instance, \methods can support TDA for non-converged models, accounting for the total number of iterations $T$ the model was trained with. It can also support TDA for multi-stage training pipelines: in a case where the model was sequentially trained with two datasets $\trainingData_1$ and $\trainingData_2$, \methods can compute the contribution of a data point $\dataPoint_m \in \trainingData_1$ that appeared in the first segment by partitioning the training trajectory into two segments ($L = 2$) and computing the expected total derivative at the first segment with $-\frac{1}{N_1} \barSegmentS_2 \bar{\mathbf{r}}_1$, where $N_1 \coloneq |\trainingData_1|$ is the size of the first training dataset.

Given terminal parameters $\params_T$ from a single training run and a query data point $\dataPoint_q$, the change in the measurable quantity due to the removal of a training data point $\dataPoint_m \in \trainingData$ can be approximated as:
\begin{align}
    f(\dataPoint_q, \params_T (-1)) - f(\dataPoint_q, \params_T (0)) \approx -\nabla_{\params} f(\dataPoint_q, \params_T)^\top \frac{\mathrm{d} \params_T}{\mathrm{d} \epsilon}.
\end{align}
Denoting $\finalParams$ as the terminal parameters trained with hyperparameters $\hyper$ and a random choice $\randomness$ (for consistency with the notations introduced in \Cref{subsec:tda}), a single-training-run estimator for \methods is defined as:
\begin{equation}    
\begin{aligned}
    \tau_{\method} (\dataPoint_q, \dataPoint_m, \trainingData; \hyperParams)  \coloneqq \nabla_{\params} f(\dataPoint_q, \finalParams)^\top \left( \sum_{\ell=1}^L \left( \prod_{\ell'=L}^{\ell+1} \bar{\segment}_{\ell'} \right) \bar{\mathbf{r}}_{\ell} \right).
    \label{eq:source_algo}
\end{aligned}
\end{equation}
Unlike the single-training-run estimator for unrolling-based approaches, \methods does not require access to the exact location where the data point $\dataPoint_m$ was used during training, as it estimates the averaged effect of removing a data point within a given segment. To further account for other sources of randomness, such as model initialization, the multiple-training-run estimator for \methods averages the final scores in \Cref{eq:source_algo} obtained for each training run with different random choices.

\subsection{Practical Algorithm for \textsc{SOURCE}}
\label{subsec:algo}

We now describe an instantiation of \methods which is practical to implement. Given the $C$ model checkpoints saved during training, \methods begins by organizing them into $L$ distinct segments. These segments may represent explicit stages in training (\eg, continual learning) or account for the change in Hessian and gradient throughout training. Within each segment $\ell$, \methods estimates the stationary Hessian $\barHess_{\ell}$ and gradient $\barGrad_{\ell}$ by averaging the Hessian and gradient across all checkpoints in the segment. When different learning rates are used within a segment, we set $\barLR_\ell$ to be the averaged learning rate, computed as $\barLR_\ell = \frac{1}{K_\ell} \sum_{k=T_{\ell - 1}}^{T_\ell - 1} \eta_k$.

However, computing \Cref{eq:unif_segment} has two practical bottlenecks for neural networks: computation of the Hessian and its matrix exponential. We fit a parametric approximation to the Hessian using Eigenvalue-corrected Kronecker-Factored Approximate Curvature (EK-FAC) \citep{george2018fast}. The EK-FAC parameterization is convenient for \methods as the approximate Hessian has an explicit eigendecomposition, which enables efficient computation of $\bar{\segment}_{\ell}$ and $\bar{\mathbf{r}}_\ell$ by applying appropriate matrix functions to the eigenvalues. Note that EK-FAC approximates the Hessian with the Gauss-Newton Hessian (GNH) \citep{martens2015optimizing}. Unlike the Hessian, the GNH is guaranteed to be positive semi-definite, as long as the loss function is convex in the model outputs \citep{martens2020new}. The GNH approximation within EK-FAC is also advantageous for \methods as it can avoid numerical instability in computing \Cref{eq:unif_segment}, especially when the Hessian has negative eigenvalues. The implementation details are provided in \Cref{app:implementation_details}.

\vspace{-0.1cm}
\paragraph{Computation Costs.} Compared to influence functions with the same EK-FAC approximation \citep{grosse2023studying}, \methods requires computing the EK-FAC factors and training gradients for each model checkpoint when performing TDA on all segments. Hence, \methods is $C$ times more computationally expensive, where $C$ is the number of checkpoints. Note that training gradients must only be computed on checkpoints within the segment when TDA is performed only on the $\ell$-th segment. In \Cref{app:avg_params}, we introduce a more computationally efficient version of \method, where we average the parameters within a segment instead of averaging Hessians and gradients. This variant of \methods is $L$ times more computationally expensive than influence functions, as the EK-FAC factors and gradients only need to be computed once for each segment.

\vspace{-0.1cm}
\paragraph{Applicability to Other Approximation Techniques.} While we described one instantiation of \methods with the EK-FAC approximation, \methods can be integrated with other techniques used for approximating implicit-differentiation-based TDA methods, such as \textsc{Trak} \citep{park2023trak} and \textsc{DataInf} \citep{kwon2023datainf}. For example, as in \textsc{Trak}, we can use random projection \citep{johnson1986extensions} to efficiently compute the averaged Hessian and gradients in a lower-dimensional space. \textsc{Trak} is advantageous over the EK-FAC approximation when there are many query data points, as it caches compressed training gradients in memory, avoiding recomputing them for each query.

\section{Related Works}
\label{sec:related_works}

Modern TDA techniques for neural networks can be broadly categorized into three main groups: sampling-based, representation-based, and gradient-based. For a comprehensive overview of TDA, including practical applications, we refer the reader to \citet{hammoudeh2022training} and \citet{mucsanyi2023trustworthy}. Sampling-based (or retraining-based) approaches, such as Shapley-value estimators \citep{shapley1953value,ghorbani2019data,jia2019towards,kwon2021beta,wang2024privacy}, \textsc{Downsampling} \citep{feldman2020neural,zhang2021counterfactual}, \textsc{Datamodels} \citep{ilyas2022datamodels}, and \textsc{Data Banzhaf} \citep{banzhaf1964weighted,wang2023data}, approximate counterfactuals by repeatedly retraining models on different data subsets. Although effective, these methods are often impractical for modern neural networks due to the significant computational cost of repeated model retraining.

Representation-based techniques evaluate the relevance between a training and query data point by examining the similarity in their representation space (\eg, the output of the last hidden layer) \citep{caruana1999case,hanawa2020evaluation}. These techniques offer computational advantages compared to other attribution methods, as they only require forward passes through the trained network. \citet{rajani2020explaining} further improves efficiency by caching all hidden representations of the training dataset and using approximate nearest neighbor search \citep{johnson2019billion}. Past works have also proposed model-agnostic TDA approaches, such as computing the similarity between query and training sequences with BM25 \citep{robertson1995okapi} for language models \citep{akyurek2022tracing,ladhak2022contrastive} or with an embedding vector obtained from a separate pre-trained self-supervised model for image classification tasks \citep{singla2023simple}. However, representation-based and input-similarity-based techniques lack a connection to the counterfactual and do not provide a notion of negatively (harmful) influential data points.

Two main strategies for gradient-based TDA are implicit differentiation and unrolling. To the best of our knowledge, the largest model to which exact unrolling has been applied is a $300$ thousand parameter model \citep{hara2019data}. Our experiments in \Cref{sec:experiments} cover TDA for models ranging from $560$ thousand parameters (MNIST \& MLP) to $120$ million parameters (WikiText-2 \& GPT-2). \textsc{SGD-Influence} \citep{hara2019data} also considers applying unrolling to only the last epoch for large-scale models. However, this limits its applicability in analyzing the effect of removing a data point at the beginning of training or analyzing multi-stage training processes. In contrast, \textsc{Hydra} \citep{chen2021hydra} approximates the mini-batch Hessian $\mathbf{H}_k$ in \Cref{eqn:unroll} as zero when computing the total derivatives, avoiding the need to compute Hessian-vector products (HVPs) for each optimization step. However, in \Cref{app:add_baseline}, we empirically observe that an accurate approximation of the Hessian is important to achieve good TDA performance. Both approaches require storing a large number of optimization variables during training. Relatedly, \citet{nickl2024memory} use local perturbation methods \citep{jaeckel1972infinitesimal} to approximate the data point's sensitivity to the training trajectory.

Apart from implicit-differentiation-based and unrolling-based approaches, \textsc{TracIn} \citep{pruthi2020estimating} is another prominent gradient-based TDA technique, which estimates the importance of a training data point by approximating the total change in the query's measurable quantity with the gradient update from this data point throughout training. Similarly to \method, the practical version of \textsc{TracIn} (\textsc{TracInCP}) leverages intermediate checkpoints saved during training. While \textsc{TracInCP} is straightforward to implement as it does not involve approximation of the Hessians, its connection to the counterfactual is unclear \citep{hammoudeh2022training,schioppa2023theoretical}. However, past works have shown its strengths in downstream tasks, such as mislabeled data detection \citep{pruthi2020estimating} and curating fine-tuning data \citep{xia2024less}.

\section{Experiments}
\label{sec:experiments}

Our experiments investigate two key questions: (1) How does \methods compare to existing TDA techniques, as measured by the linear datamodeling score (LDS) and through subset removal counterfactual evaluation? (2) Can \methods support data attribution in situations where implicit-differentiation-based approaches struggle, particularly with models that have not converged or have been trained in multiple stages with different datasets?

\subsection{Experimental Setup}
\label{subsec:exp_setup}

Our experiments consider diverse machine learning tasks, including: (a) regression using datasets from the UCI Machine Learning Repository \citep{dua2019uci}, (b) image classification with datasets such as MNIST \citep{lecun2010mnist}, FashionMNIST \citep{xiao2017fashion}, CIFAR-10 \citep{krizhevsky2009learning}, RotatedMNIST \citep{ghifary2015domain}, and PACS \citep{li2017deeper}, (c) text classification using the GLUE benchmark \citep{wang2018glue}, and (d) language modeling with the WikiText-2 dataset \citep{merity2016pointer}. A detailed description of each task is provided in \Cref{app:experiment_details_datasets_models}.

Across these tasks, we compare \methods against existing TDA techniques: representation similarity (\textsc{RepSim}) \citep{caruana1999case,hanawa2020evaluation}, \textsc{TracIn} \citep{pruthi2020estimating}, \textsc{Trak} \citep{park2023trak} and influence functions (\textsc{IF}) with the EK-FAC approximation \citep{grosse2023studying}.\footnote{In \Cref{app:add_baseline}, we also include empirical influence (\textsc{Downsampling}) \citep{feldman2020neural} and \textsc{Hydra} \citep{chen2021hydra} as baselines for the FashionMNIST task. These baselines were omitted for other tasks due to the large computational costs involved.} The implementation details of our baseline techniques are provided in \Cref{app:baseline}. For consistency with \citet{park2023trak}, the measurement $\measurement$ is defined as the margin for classification tasks and the absolute error for regression tasks. We set the measurement as the loss for language modeling.

Our evaluations are conducted under two separate settings. First is a single model setup, where TDA techniques use model checkpoints from a single training run. Unless specified otherwise, \textsc{RepSim}, \textsc{Trak}, and \textsc{IF} are computed at the final training checkpoint, and \textsc{TracIn} and \methods use $6$ checkpoints saved throughout training. \methods use $3$ segments ($L = 3$) equally partitioned at the early, middle, and late stages of training. In the second setting, TDA techniques use checkpoints from $10$ distinct models, each trained with varying sources of randomness. Past works have shown ensembling attribution scores across models can improve TDA performance \citep{park2023trak,nguyen2023bayesian}. For all TDA techniques, including \method, we simply average the final attribution scores from distinctly trained models with the full dataset, except for \textsc{Trak}, which uses its custom ensembling procedures with models trained on $50\%$ of the original dataset.

\begin{figure*}[!t]
    \centering
    \vspace{-0.2cm}
    \includegraphics{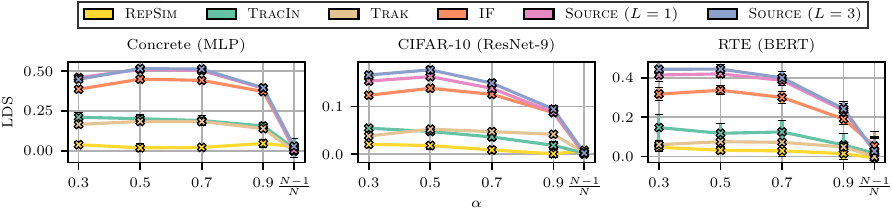}
    \vspace{-0.25cm}
    \caption{Linear datamodeling scores (LDS) across a range of data sampling ratios $\alpha$ for \methods ($L = \{1, 3\}$) and baseline TDA techniques. The LDS is measured for a single model setup, and error bars represent $95\%$ bootstrap confidence intervals.}
    \label{fig:each_lds}
\end{figure*}

\subsection{Evaluations with Linear Datamodeling Score (LDS)}
\label{subsec:lds_eval}

We first consider computing the linear datamodeling score (LDS), defined in \Cref{subsec:tda_evaluate}, across a range of data sampling ratios $\alpha$. (The procedures to compute the LDS are described in \Cref{app:lds}.) The performance of \methods and other baseline attribution methods is shown in \Cref{fig:each_lds}. \methods consistently achieves higher LDS than the baseline methods across diverse $\alpha$ values. However, an exception is noted at $\alpha = 1 - \sfrac{1}{N}$ (\eg, removing a single training data point), where a significant drop in correlations is observed for all TDA methods. This finding is consistent with previous studies that highlight the limitations of \loos estimates in reliably evaluating attribution techniques \citep{sogaard2021revisiting,epifano2023revisiting,nguyen2023bayesian} (see \Cref{app:loo} for a detailed discussion). Additionally, our results suggest that while \methods with a single segment can be effective, using multiple segments typically improves LDS performance.

\begin{figure*}[!t]
    \vspace{-0.8cm}
    \centering
    \includegraphics{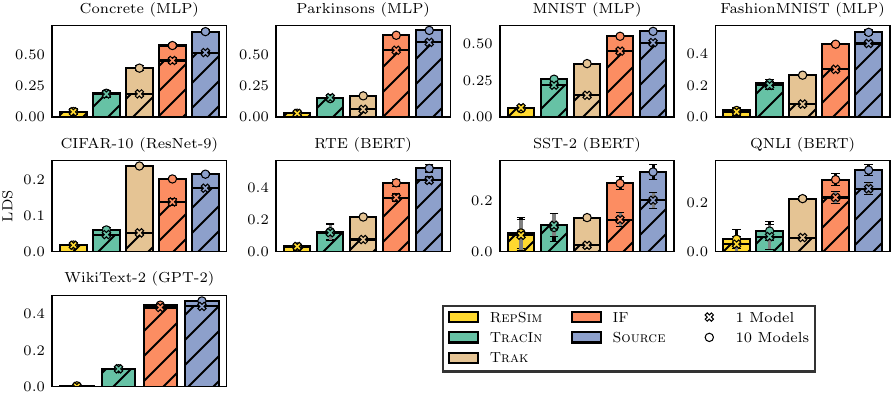}
    \vspace{-0.25cm}
    \caption{Linear datamodeling scores (LDS) at $\alpha = 0.5$ for \methods ($L = 3$) and baseline TDA techniques on regression, image classification, text classification, and language modeling tasks. The error bars represent $95\%$ bootstrap confidence intervals. (Results for \textsc{Trak} on WikiText-2 are omitted due to the lack of publicly available implementations for language modeling tasks.)}
    \label{fig:nn_lds}
\end{figure*}

Given that the relative rankings of TDA techniques typically remain consistent across various $\alpha$ values, we present the LDS results at $\alpha = 0.5$ for additional tasks in \Cref{fig:nn_lds}. \method\ consistently outperforms baseline methods in a single model setup, achieving higher correlations with the ground truth. When aggregating TDA scores from multiple models, we observe a large improvement in the LDS, particularly for \textsc{Trak}, \textsc{IF}, and \method. \method\ achieves the highest LDS across all tasks, except for the CIFAR-10 classification task using ResNet-9. However, we show that \methods outperforms baseline methods on the CIFAR-10 task for subset removal counterfactual evaluation in \Cref{subsec:applications}.

\begin{figure*}[!t]
    \centering
    \vspace{-0.2cm}
    \includegraphics{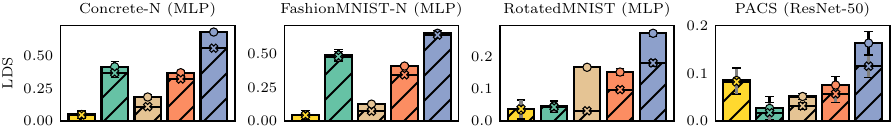}
    \vspace{-0.25cm}
    \caption{Linear datamodeling scores (LDS) at $\alpha = 0.5$ for \methods and baseline TDA techniques on settings that pose challenges to implicit-differentiation-based TDA techniques (\eg, influence functions). See \Cref{subsec:failure_exp} for a detailed description of these settings and \Cref{fig:nn_lds} for labels.}
    \label{fig:other_lds}
\end{figure*}

\subsection{TDA Evaluations on Other Training Scenarios}
\label{subsec:failure_exp}

In \Cref{subsec:lds_eval}, we considered models that are sufficiently trained near convergence using a fixed dataset, where implicit-differentiation-based methods are expected to perform similarly to unrolling-based methods. We now investigate two scenarios that pose challenges for implicit-differentiation-based TDA techniques. These are: (1) non-converged models trained with only a small number of update iterations and (2) models trained sequentially with two distinct datasets, a common setup in continual learning. We demonstrate that \methods offers distinct advantages over implicit-differentiation-based approaches in these contexts. The effectiveness of \methods in these scenarios, as measured by the LDS, is shown in \Cref{fig:other_lds}. \method\ performs strongly against other baseline techniques in these setups, and indeed, even the non-ensembled version of \method\ typically outperforms the ensembled versions of the competing methods.

\vspace{-0.1cm}
\paragraph{TDA for Non-Converged Models.} In our first scenario, we assess the effectiveness of TDA techniques for models trained with a small number of update steps. We use versions of the Concrete and FashionMNIST datasets that have been modified -- either by corrupting target values or relabeling $30\%$ of the data points. Then, we train the networks for only $3$ epochs to avoid overfitting. We use $3$ intermediate checkpoints (at the end of each epoch) for \textsc{TracIn} and \method. On both tasks, influence functions are less effective than \textsc{TracIn} (despite having performed better in the previous experiments). However, \methods still achieves the best performance, consistent with its non-reliance on the optimality of the final weights. In \Cref{app:counterfactual_linear}, we show that this observation -- that \methods outperforms influence functions for models that have not fully converged -- also holds for linear models.

\vspace{-0.1cm}
\paragraph{TDA for Sequentially Trained Models.} In many practical applications, networks are trained sequentially, each phase using different datasets or objectives. We consider a setup where a model is initially trained with a dataset $\trainingData_1$, and subsequently trained with another dataset $\trainingData_2$. We use test examples from $\trainingData_2$ for query data points and attribute the final model's behavior to the first dataset. In other words, we aim to investigate the impact of removing training data points in the first training stage on the final model behavior (further trained on another dataset). This is a more challenging setting, as sequential training has shown catastrophic forgetting \citep{goodfellow2013empirical}. Since implicit-differentiation-based methods such as \textsc{Trak} and \textsc{IF} do not provide any way to separate multiple stages of training, for these methods, we simply combine the data from both stages into a larger dataset for TDA. We use two segments for \methods ($L = 2$), partitioned at different stages, and perform TDA only for the first segment.

Our experiments use the RotatedMNIST and PACS datasets, both containing multiple data distributions. For example, RotatedMNIST contains five unique domains differentiated by the rotation angles of the images: $0$, $15$, $30$, $45$, and $60$ degrees. We select one of these domains for the second retraining stage, while the remaining domains are used in the first training stage. Similarly to the non-converged settings, \methods performs strongly against other baseline techniques in the continual learning settings.

\subsection{Subset Removal Counterfactual Evaluation}
\label{subsec:applications}

\begin{figure*}[!t]
    \centering
    \includegraphics{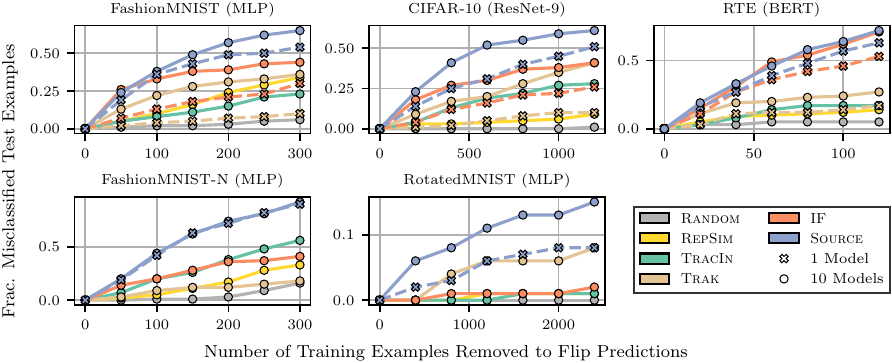}
    \vspace{-0.25cm}
    \caption{Subset removal counterfactual evaluation for \methods and baseline TDA techniques, where the top positively influential data points predicted by each TDA method are removed, and the model is retrained to misclassify a (previously correctly classified) test data point.}
    \label{fig:brittleness}
\end{figure*}

So far, we have focused on quantifying TDA accuracy using the LDS. Another approach to assess the effectiveness of TDA techniques is subset removal counterfactual evaluation (see \Cref{subsec:tda_evaluate}), which examines the change in model behavior before and after removing data points highly ranked in influence by different attribution techniques. Effective data attribution methods should identify data points whose exclusions lead to significant changes in model behavior.

We considered FashionMNIST, CIFAR-10, RTE, and RotatedMNIST classification tasks from \Cref{subsec:lds_eval} and \Cref{subsec:failure_exp}. We first selected $100$ test examples that were initially correctly classified (across all $5$ random seeds) when trained with the entire dataset $\trainingData$. Then, for each test example $\dataPoint_q$ and TDA technique, we identified the top-$k$ most positively influential training data points, removed these data points from the original dataset, and retrained the model with this modified dataset. We report the fraction of test examples (out of the selected $100$ test points) that get misclassified on average (over $3$ random seeds) after removing at most $k$ positively influential training data points. (The detailed procedures are described in \Cref{app:counter}.) The results are shown in \Cref{fig:brittleness}. We observe that \methods better identifies the top influential data points causing misclassification than other baseline TDA techniques. The improvement is more substantial for settings in \Cref{subsec:failure_exp} that pose challenges to implicit-differentiation-based approaches.

\section{Conclusion}

We introduced \methods (\textbf{S}egmented stati\textbf{O}nary \textbf{U}n\textbf{R}olling for \textbf{C}ounterfactual \textbf{E}stimation), a novel TDA technique that combines the strengths of implicit-differentiation-based and unrolling-based techniques. \methods approximates unrolled differentiation by partitioning the training trajectory into one or more segments and approximating the gradients and Hessians as stationary within each segment, yielding an influence-function-like estimation algorithm. We showed one instantiation of \methods by approximating the Hessian with the EK-FAC parameterization. On a diverse task set, we demonstrated \method's effectiveness compared to existing data attribution techniques, especially when the network has not converged or has been trained with multiple stages.

\section*{Acknowledgements}

The authors would like to thank Jenny Bao, Rob Brekelmans, Sang Keun Choe, Lev McKinney, Andrew Wang, and Arielle Zhang for their helpful feedback on the manuscript. Resources used in preparing this research were provided, in part, by the Province of Ontario, the Government of Canada through CIFAR, and companies sponsoring the Vector Institute: \url{www.vectorinstitute.ai/#partners}. JB was funded by OpenPhilanthropy and Good Ventures. RG acknowledges support from the Canada CIFAR AI Chairs program.

\bibliography{bibliography}

\begin{thebibliography}{99}
\providecommand{\natexlab}[1]{#1}
\providecommand{\url}[1]{\texttt{#1}}
\expandafter\ifx\csname urlstyle\endcsname\relax
  \providecommand{\doi}[1]{doi: #1}\else
  \providecommand{\doi}{doi: \begingroup \urlstyle{rm}\Url}\fi

\bibitem[Aky{\"u}rek et~al.(2022)Aky{\"u}rek, Bolukbasi, Liu, Xiong, Tenney, Andreas, and Guu]{akyurek2022tracing}
Ekin Aky{\"u}rek, Tolga Bolukbasi, Frederick Liu, Binbin Xiong, Ian Tenney, Jacob Andreas, and Kelvin Guu.
\newblock Towards tracing knowledge in language models back to the training data.
\newblock In \emph{Findings of the Association for Computational Linguistics: EMNLP 2022}, pages 2429--2446, 2022.

\bibitem[Arnoldi(1951)]{arnoldi1951principle}
Walter~Edwin Arnoldi.
\newblock The principle of minimized iterations in the solution of the matrix eigenvalue problem.
\newblock \emph{Quarterly of applied mathematics}, 9\penalty0 (1):\penalty0 17--29, 1951.

\bibitem[Bae et~al.(2022{\natexlab{a}})Bae, Ng, Lo, Ghassemi, and Grosse]{bae2022if}
Juhan Bae, Nathan Ng, Alston Lo, Marzyeh Ghassemi, and Roger~B Grosse.
\newblock If influence functions are the answer, then what is the question?
\newblock \emph{Advances in Neural Information Processing Systems}, 35:\penalty0 17953--17967, 2022{\natexlab{a}}.

\bibitem[Bae et~al.(2022{\natexlab{b}})Bae, Vicol, HaoChen, and Grosse]{bae2022amortized}
Juhan Bae, Paul Vicol, Jeff~Z HaoChen, and Roger~B Grosse.
\newblock Amortized proximal optimization.
\newblock \emph{Advances in Neural Information Processing Systems}, 35:\penalty0 8982--8997, 2022{\natexlab{b}}.

\bibitem[Banzhaf~III(1964)]{banzhaf1964weighted}
John~F Banzhaf~III.
\newblock Weighted voting doesn't work: {A} mathematical analysis.
\newblock \emph{Rutgers L. Rev.}, 19:\penalty0 317, 1964.

\bibitem[Basu et~al.(2020)Basu, Pope, and Feizi]{basu2020influence}
Samyadeep Basu, Phil Pope, and Soheil Feizi.
\newblock Influence functions in deep learning are fragile.
\newblock In \emph{International Conference on Learning Representations}, 2020.

\bibitem[Bengio(2012)]{bengio2012practical}
Yoshua Bengio.
\newblock Practical recommendations for gradient-based training of deep architectures.
\newblock In \emph{Neural Networks: Tricks of the Trade: Second Edition}, pages 437--478. Springer, 2012.

\bibitem[Brophy et~al.(2023)Brophy, Hammoudeh, and Lowd]{brophy2023adapting}
Jonathan Brophy, Zayd Hammoudeh, and Daniel Lowd.
\newblock Adapting and evaluating influence-estimation methods for gradient-boosted decision trees.
\newblock \emph{Journal of Machine Learning Research}, 24\penalty0 (154):\penalty0 1--48, 2023.

\bibitem[Caruana et~al.(1999)Caruana, Kangarloo, Dionisio, Sinha, and Johnson]{caruana1999case}
Rich Caruana, Hooshang Kangarloo, John~David Dionisio, Usha Sinha, and David Johnson.
\newblock Case-based explanation of non-case-based learning methods.
\newblock In \emph{Proceedings of the AMIA Symposium}, page 212. American Medical Informatics Association, 1999.

\bibitem[Chen et~al.(2021)Chen, Li, Yu, Wu, and Miao]{chen2021hydra}
Yuanyuan Chen, Boyang Li, Han Yu, Pengcheng Wu, and Chunyan Miao.
\newblock Hydra: {Hypergradient} data relevance analysis for interpreting deep neural networks.
\newblock In \emph{Proceedings of the AAAI Conference on Artificial Intelligence}, volume~35, pages 7081--7089, 2021.

\bibitem[Devlin et~al.(2018)Devlin, Chang, Lee, and Toutanova]{devlin2018bert}
Jacob Devlin, Ming-Wei Chang, Kenton Lee, and Kristina Toutanova.
\newblock {BERT}: {Pre}-training of deep bidirectional transformers for language understanding, 2018.

\bibitem[Engstrom et~al.(2024)Engstrom, Feldmann, and Madry]{engstrom2024dsdm}
Logan Engstrom, Axel Feldmann, and Aleksander Madry.
\newblock {DsDm}: {Model}-aware dataset selection with datamodels, 2024.

\bibitem[Epifano et~al.(2023)Epifano, Ramachandran, Masino, and Rasool]{epifano2023revisiting}
Jacob~R Epifano, Ravi~P Ramachandran, Aaron~J Masino, and Ghulam Rasool.
\newblock Revisiting the fragility of influence functions.
\newblock \emph{Neural Networks}, 162:\penalty0 581--588, 2023.

\bibitem[Eschenhagen et~al.(2024)Eschenhagen, Immer, Turner, Schneider, and Hennig]{eschenhagen2024kronecker}
Runa Eschenhagen, Alexander Immer, Richard Turner, Frank Schneider, and Philipp Hennig.
\newblock Kronecker-factored approximate curvature for modern neural network architectures.
\newblock \emph{Advances in Neural Information Processing Systems}, 36, 2024.

\bibitem[Fang et~al.(2020)Fang, Gong, and Liu]{fang2020influence}
Minghong Fang, Neil~Zhenqiang Gong, and Jia Liu.
\newblock Influence function based data poisoning attacks to top-$n$ recommender systems.
\newblock In \emph{Proceedings of The Web Conference 2020}, pages 3019--3025, 2020.

\bibitem[Feldman and Zhang(2020)]{feldman2020neural}
Vitaly Feldman and Chiyuan Zhang.
\newblock What neural networks memorize and why: Discovering the long tail via influence estimation.
\newblock \emph{Advances in Neural Information Processing Systems}, 33:\penalty0 2881--2891, 2020.

\bibitem[George et~al.(2018)George, Laurent, Bouthillier, Ballas, and Vincent]{george2018fast}
Thomas George, C{\'e}sar Laurent, Xavier Bouthillier, Nicolas Ballas, and Pascal Vincent.
\newblock Fast approximate natural gradient descent in a kronecker factored eigenbasis.
\newblock \emph{Advances in Neural Information Processing Systems}, 31, 2018.

\bibitem[Georgiev et~al.(2023)Georgiev, Vendrow, Salman, Park, and Madry]{georgiev2023journey}
Kristian Georgiev, Joshua Vendrow, Hadi Salman, Sung~Min Park, and Aleksander Madry.
\newblock The journey, not the destination: {How} data guides diffusion models, 2023.

\bibitem[Ghifary et~al.(2015)Ghifary, Kleijn, Zhang, and Balduzzi]{ghifary2015domain}
Muhammad Ghifary, W~Bastiaan Kleijn, Mengjie Zhang, and David Balduzzi.
\newblock Domain generalization for object recognition with multi-task autoencoders.
\newblock In \emph{Proceedings of the IEEE international conference on computer vision}, pages 2551--2559, 2015.

\bibitem[Ghorbani and Zou(2019)]{ghorbani2019data}
Amirata Ghorbani and James Zou.
\newblock {Data Shapley: Equitable} valuation of data for machine learning.
\newblock In \emph{International Conference on Machine Learning}, pages 2242--2251. PMLR, 2019.

\bibitem[Goodfellow et~al.(2015)Goodfellow, Mirza, Xiao, Courville, and Bengio]{goodfellow2013empirical}
Ian~J. Goodfellow, Mehdi Mirza, Da~Xiao, Aaron Courville, and Yoshua Bengio.
\newblock An empirical investigation of catastrophic forgetting in gradient-based neural networks, 2015.

\bibitem[Grosse(2021)]{GrosseNNTDChapter4}
Roger Grosse.
\newblock {University of Toronto CSC2541, Topics in Machine Learning: Neural Net Training Dynamics, Chapter 4: Second-Order Optimization}.
\newblock {Lecture Notes}, 2021.
\newblock URL \url{https://www.cs.toronto.edu/~rgrosse/courses/csc2541_2021/readings/L04_second_order.pdf}.

\bibitem[Grosse and Martens(2016)]{grosse2016kronecker}
Roger Grosse and James Martens.
\newblock A kronecker-factored approximate fisher matrix for convolution layers.
\newblock In \emph{International Conference on Machine Learning}, pages 573--582. PMLR, 2016.

\bibitem[Grosse et~al.(2023)Grosse, Bae, Anil, Elhage, Tamkin, Tajdini, Steiner, Li, Durmus, Perez, Hubinger, Lukošiūtė, Nguyen, Joseph, McCandlish, Kaplan, and Bowman]{grosse2023studying}
Roger Grosse, Juhan Bae, Cem Anil, Nelson Elhage, Alex Tamkin, Amirhossein Tajdini, Benoit Steiner, Dustin Li, Esin Durmus, Ethan Perez, Evan Hubinger, Kamilė Lukošiūtė, Karina Nguyen, Nicholas Joseph, Sam McCandlish, Jared Kaplan, and Samuel~R. Bowman.
\newblock Studying large language model generalization with influence functions, 2023.

\bibitem[Gulrajani and Lopez-Paz(2020)]{gulrajani2020search}
Ishaan Gulrajani and David Lopez-Paz.
\newblock In search of lost domain generalization, 2020.

\bibitem[Gupta et~al.(2018)Gupta, Koren, and Singer]{gupta2018shampoo}
Vineet Gupta, Tomer Koren, and Yoram Singer.
\newblock {Shampoo}: {Preconditioned} stochastic tensor optimization.
\newblock In \emph{International Conference on Machine Learning}, pages 1842--1850. PMLR, 2018.

\bibitem[Guu et~al.(2023)Guu, Webson, Pavlick, Dixon, Tenney, and Bolukbasi]{guu2023simfluence}
Kelvin Guu, Albert Webson, Ellie Pavlick, Lucas Dixon, Ian Tenney, and Tolga Bolukbasi.
\newblock Simfluence: Modeling the influence of individual training examples by simulating training runs, 2023.

\bibitem[Hammoudeh and Lowd(2024)]{hammoudeh2022training}
Zayd Hammoudeh and Daniel Lowd.
\newblock Training data influence analysis and estimation: {A} survey.
\newblock \emph{Machine Learning}, pages 1--53, 2024.

\bibitem[Hampel(1974)]{hampel1974influence}
Frank~R Hampel.
\newblock The influence curve and its role in robust estimation.
\newblock \emph{Journal of the american statistical association}, 69\penalty0 (346):\penalty0 383--393, 1974.

\bibitem[Han et~al.(2020)Han, Wallace, and Tsvetkov]{han2020explaining}
Xiaochuang Han, Byron~C Wallace, and Yulia Tsvetkov.
\newblock Explaining black box predictions and unveiling data artifacts through influence functions.
\newblock In \emph{Proceedings of the 58th Annual Meeting of the Association for Computational Linguistics}, pages 5553--5563, 2020.

\bibitem[Hanawa et~al.(2020)Hanawa, Yokoi, Hara, and Inui]{hanawa2020evaluation}
Kazuaki Hanawa, Sho Yokoi, Satoshi Hara, and Kentaro Inui.
\newblock Evaluation of similarity-based explanations.
\newblock In \emph{International Conference on Learning Representations}, 2020.

\bibitem[Hara et~al.(2019)Hara, Nitanda, and Maehara]{hara2019data}
Satoshi Hara, Atsushi Nitanda, and Takanori Maehara.
\newblock Data cleansing for models trained with sgd.
\newblock \emph{Advances in Neural Information Processing Systems}, 32, 2019.

\bibitem[He et~al.(2016)He, Zhang, Ren, and Sun]{he2016deep}
Kaiming He, Xiangyu Zhang, Shaoqing Ren, and Jian Sun.
\newblock Deep residual learning for image recognition.
\newblock In \emph{Proceedings of the IEEE conference on computer vision and pattern recognition}, pages 770--778, 2016.

\bibitem[Hooker et~al.(2019)Hooker, Erhan, Kindermans, and Kim]{hooker2019benchmark}
Sara Hooker, Dumitru Erhan, Pieter-Jan Kindermans, and Been Kim.
\newblock A benchmark for interpretability methods in deep neural networks.
\newblock \emph{Advances in neural information processing systems}, 32, 2019.

\bibitem[Hu et~al.(2021)Hu, Shen, Wallis, Allen-Zhu, Li, Wang, Wang, and Chen]{hu2021lora}
Edward~J. Hu, Yelong Shen, Phillip Wallis, Zeyuan Allen-Zhu, Yuanzhi Li, Shean Wang, Lu~Wang, and Weizhu Chen.
\newblock Lora: Low-rank adaptation of large language models, 2021.

\bibitem[Ilyas et~al.(2022)Ilyas, Park, Engstrom, Leclerc, and Madry]{ilyas2022datamodels}
Andrew Ilyas, Sung~Min Park, Logan Engstrom, Guillaume Leclerc, and Aleksander Madry.
\newblock {Datamodels}: {Predicting} predictions from training data.
\newblock In \emph{International Conference on Machine Learning}, 2022.

\bibitem[Izadi et~al.(2020)Izadi, Fang, Stevenson, and Lin]{izadi2020optimization}
Mohammad~Rasool Izadi, Yihao Fang, Robert Stevenson, and Lizhen Lin.
\newblock Optimization of graph neural networks with natural gradient descent.
\newblock In \emph{2020 IEEE international conference on big data}, pages 171--179. IEEE, 2020.

\bibitem[Jaeckel(1972)]{jaeckel1972infinitesimal}
Louis~A Jaeckel.
\newblock \emph{The infinitesimal jackknife}.
\newblock Bell Telephone Laboratories, 1972.

\bibitem[Jagielski et~al.(2021)Jagielski, Severi, Pousette~Harger, and Oprea]{jagielski2021subpopulation}
Matthew Jagielski, Giorgio Severi, Niklas Pousette~Harger, and Alina Oprea.
\newblock Subpopulation data poisoning attacks.
\newblock In \emph{Proceedings of the 2021 ACM SIGSAC Conference on Computer and Communications Security}, pages 3104--3122, 2021.

\bibitem[Jia et~al.(2019)Jia, Dao, Wang, Hubis, Hynes, G{\"u}rel, Li, Zhang, Song, and Spanos]{jia2019towards}
Ruoxi Jia, David Dao, Boxin Wang, Frances~Ann Hubis, Nick Hynes, Nezihe~Merve G{\"u}rel, Bo~Li, Ce~Zhang, Dawn Song, and Costas~J Spanos.
\newblock Towards efficient data valuation based on the shapley value.
\newblock In \emph{The 22nd International Conference on Artificial Intelligence and Statistics}, pages 1167--1176. PMLR, 2019.

\bibitem[Jia et~al.(2021)Jia, Wu, Sun, Xu, Dao, Kailkhura, Zhang, Li, and Song]{jia2021scalability}
Ruoxi Jia, Fan Wu, Xuehui Sun, Jiacen Xu, David Dao, Bhavya Kailkhura, Ce~Zhang, Bo~Li, and Dawn Song.
\newblock Scalability vs. utility: Do we have to sacrifice one for the other in data importance quantification?
\newblock In \emph{Proceedings of the IEEE/CVF Conference on Computer Vision and Pattern Recognition}, pages 8239--8247, 2021.

\bibitem[Jiang et~al.(2023)Jiang, Liang, Zou, and Kwon]{jiang2023opendataval}
Kevin~Fu Jiang, Weixin Liang, James Zou, and Yongchan Kwon.
\newblock Opendataval: {A} unified benchmark for data valuation, 2023.

\bibitem[Johnson et~al.(2019)Johnson, Douze, and J{\'e}gou]{johnson2019billion}
Jeff Johnson, Matthijs Douze, and Herv{\'e} J{\'e}gou.
\newblock Billion-scale similarity search with {GPUs}.
\newblock \emph{IEEE Transactions on Big Data}, 7\penalty0 (3):\penalty0 535--547, 2019.

\bibitem[Johnson et~al.(1986)Johnson, Lindenstrauss, and Schechtman]{johnson1986extensions}
William~B Johnson, Joram Lindenstrauss, and Gideon Schechtman.
\newblock Extensions of lipschitz maps into banach spaces.
\newblock \emph{Israel Journal of Mathematics}, 54\penalty0 (2):\penalty0 129--138, 1986.

\bibitem[K and Søgaard(2021)]{sogaard2021revisiting}
Karthikeyan K and Anders Søgaard.
\newblock Revisiting methods for finding influential examples, 2021.

\bibitem[Kelly et~al.(2023)Kelly, Longjohn, and Nottingham]{dua2019uci}
Markelle Kelly, Rachel Longjohn, and Kolby Nottingham.
\newblock The {UCI} machine learning repository, 2023.
\newblock URL \url{https://archive.ics.uci.edu}.

\bibitem[Khanna et~al.(2019)Khanna, Kim, Ghosh, and Koyejo]{khanna2019interpreting}
Rajiv Khanna, Been Kim, Joydeep Ghosh, and Sanmi Koyejo.
\newblock Interpreting black box predictions using fisher kernels.
\newblock In \emph{The 22nd International Conference on Artificial Intelligence and Statistics}, pages 3382--3390. PMLR, 2019.

\bibitem[Kim et~al.(2024)Kim, Kim, and Yang]{kim2023gex}
SungYub Kim, Kyungsu Kim, and Eunho Yang.
\newblock {GEX}: {A} flexible method for approximating influence via geometric ensemble.
\newblock \emph{Advances in Neural Information Processing Systems}, 36, 2024.

\bibitem[Kingma and Ba(2014)]{kingma2014adam}
Diederik~P. Kingma and Jimmy Ba.
\newblock Adam: A method for stochastic optimization, 2014.

\bibitem[Koh and Liang(2017)]{koh2017understanding}
Pang~Wei Koh and Percy Liang.
\newblock Understanding black-box predictions via influence functions.
\newblock In \emph{International Conference on Machine Learning}, pages 1885--1894. PMLR, 2017.

\bibitem[Kong et~al.(2021)Kong, Shen, and Huang]{kong2021resolving}
Shuming Kong, Yanyan Shen, and Linpeng Huang.
\newblock Resolving training biases via influence-based data relabeling.
\newblock In \emph{International Conference on Learning Representations}, 2021.

\bibitem[Konz et~al.(2023)Konz, Godfrey, Shapiro, Tu, Kvinge, and Brown]{konz2023attributing}
Nicholas Konz, Charles Godfrey, Madelyn Shapiro, Jonathan Tu, Henry Kvinge, and Davis Brown.
\newblock Attributing learned concepts in neural networks to training data, 2023.

\bibitem[Krantz and Parks(2002)]{krantz2002implicit}
Steven~George Krantz and Harold~R Parks.
\newblock \emph{{The Implicit Function Theorem}: {History}, theory, and applications}.
\newblock Springer Science \& Business Media, 2002.

\bibitem[Krizhevsky and Hinton(2009)]{krizhevsky2009learning}
Alex Krizhevsky and Geoffrey Hinton.
\newblock Learning multiple layers of features from tiny images.
\newblock 2009.

\bibitem[Kwon and Zou(2022)]{kwon2021beta}
Yongchan Kwon and James Zou.
\newblock {Beta Shapley}: {A} unified and noise-reduced data valuation framework for machine learning.
\newblock In \emph{International Conference on Artificial Intelligence and Statistics}, pages 8780--8802. PMLR, 2022.

\bibitem[Kwon et~al.(2023)Kwon, Wu, Wu, and Zou]{kwon2023datainf}
Yongchan Kwon, Eric Wu, Kevin Wu, and James Zou.
\newblock {DataInf}: {Efficiently} estimating data influence in lora-tuned llms and diffusion models.
\newblock In \emph{International Conference on Learning Representations}, 2023.

\bibitem[Ladhak et~al.(2023)Ladhak, Durmus, and Hashimoto]{ladhak2022contrastive}
Faisal Ladhak, Esin Durmus, and Tatsunori~B Hashimoto.
\newblock Contrastive error attribution for finetuned language models.
\newblock In \emph{Proceedings of the 61st Annual Meeting of the Association for Computational Linguistics (Volume 1: Long Papers)}, pages 11482--11498, 2023.

\bibitem[LeCun et~al.(2010)LeCun, Cortes, and Burges]{lecun2010mnist}
Yann LeCun, Corinna Cortes, and CJ~Burges.
\newblock {MNIST} handwritten digit database.
\newblock \emph{ATT Labs}, 2, 2010.

\bibitem[Li et~al.(2017)Li, Yang, Song, and Hospedales]{li2017deeper}
Da~Li, Yongxin Yang, Yi-Zhe Song, and Timothy~M Hospedales.
\newblock Deeper, broader and artier domain generalization.
\newblock In \emph{Proceedings of the IEEE international conference on computer vision}, pages 5542--5550, 2017.

\bibitem[Liu and Nocedal(1989)]{liu1989limited}
Dong~C Liu and Jorge Nocedal.
\newblock On the limited memory {BFGS} method for large scale optimization.
\newblock \emph{Mathematical programming}, 45\penalty0 (1-3):\penalty0 503--528, 1989.

\bibitem[Liu et~al.(2021)Liu, Ding, Zhong, Li, Dai, and He]{liu2021influence}
Zhuoming Liu, Hao Ding, Huaping Zhong, Weijia Li, Jifeng Dai, and Conghui He.
\newblock Influence selection for active learning.
\newblock In \emph{Proceedings of the IEEE/CVF International Conference on Computer Vision}, pages 9274--9283, 2021.

\bibitem[Loshchilov and Hutter(2018)]{Loshchilov2017decoupled}
Ilya Loshchilov and Frank Hutter.
\newblock Decoupled weight decay regularization.
\newblock In \emph{International Conference on Learning Representations}, 2018.

\bibitem[Martens(2020)]{martens2020new}
James Martens.
\newblock New insights and perspectives on the natural gradient method.
\newblock \emph{Journal of Machine Learning Research}, 21\penalty0 (146):\penalty0 1--76, 2020.

\bibitem[Martens and Grosse(2015)]{martens2015optimizing}
James Martens and Roger Grosse.
\newblock Optimizing neural networks with kronecker-factored approximate curvature.
\newblock In \emph{International Conference on Machine Learning}, pages 2408--2417. PMLR, 2015.

\bibitem[Martens et~al.(2018)Martens, Ba, and Johnson]{martens2018kronecker}
James Martens, Jimmy Ba, and Matt Johnson.
\newblock Kronecker-factored curvature approximations for recurrent neural networks.
\newblock In \emph{International Conference on Learning Representations}, 2018.

\bibitem[Merity et~al.(2016)Merity, Xiong, Bradbury, and Socher]{merity2016pointer}
Stephen Merity, Caiming Xiong, James Bradbury, and Richard Socher.
\newblock Pointer sentinel mixture models.
\newblock In \emph{International Conference on Learning Representations}, 2016.

\bibitem[Mucsányi et~al.(2023)Mucsányi, Kirchhof, Nguyen, Rubinstein, and Oh]{mucsanyi2023trustworthy}
Bálint Mucsányi, Michael Kirchhof, Elisa Nguyen, Alexander Rubinstein, and Seong~Joon Oh.
\newblock Trustworthy machine learning, 2023.

\bibitem[Nguyen et~al.(2024)Nguyen, Seo, and Oh]{nguyen2023bayesian}
Elisa Nguyen, Minjoon Seo, and Seong~Joon Oh.
\newblock A bayesian approach to analysing training data attribution in deep learning.
\newblock \emph{Advances in Neural Information Processing Systems}, 36, 2024.

\bibitem[Nickl et~al.(2024)Nickl, Xu, Tailor, M{\"o}llenhoff, and Khan]{nickl2024memory}
Peter Nickl, Lu~Xu, Dharmesh Tailor, Thomas M{\"o}llenhoff, and Mohammad Emtiyaz~E Khan.
\newblock The memory-perturbation equation: Understanding model's sensitivity to data.
\newblock \emph{Advances in Neural Information Processing Systems}, 36, 2024.

\bibitem[Oh et~al.(2022)Oh, Ustun, McAuley, and Kumar]{oh2022rank}
Sejoon Oh, Berk Ustun, Julian McAuley, and Srijan Kumar.
\newblock Rank list sensitivity of recommender systems to interaction perturbations.
\newblock In \emph{Proceedings of the 31st ACM International Conference on Information \& Knowledge Management}, pages 1584--1594, 2022.

\bibitem[Park et~al.(2023)Park, Georgiev, Ilyas, Leclerc, and Madry]{park2023trak}
Sung~Min Park, Kristian Georgiev, Andrew Ilyas, Guillaume Leclerc, and Aleksander Madry.
\newblock {TRAK}: {Attributing} model behavior at scale.
\newblock In \emph{International Conference on Machine Learning}, pages 27074--27113. PMLR, 2023.

\bibitem[Paszke et~al.(2017)Paszke, Gross, Chintala, Chanan, Yang, DeVito, Lin, Desmaison, Antiga, and Lerer]{paszke2017automatic}
Adam Paszke, Sam Gross, Soumith Chintala, Gregory Chanan, Edward Yang, Zachary DeVito, Zeming Lin, Alban Desmaison, Luca Antiga, and Adam Lerer.
\newblock Automatic differentiation in {PyTorch}.
\newblock 2017.

\bibitem[Pruthi et~al.(2020)Pruthi, Liu, Kale, and Sundararajan]{pruthi2020estimating}
Garima Pruthi, Frederick Liu, Satyen Kale, and Mukund Sundararajan.
\newblock Estimating training data influence by tracing gradient descent.
\newblock \emph{Advances in Neural Information Processing Systems}, 33:\penalty0 19920--19930, 2020.

\bibitem[Radford et~al.(2019)Radford, Wu, Child, Luan, Amodei, and Sutskever]{radford2019language}
Alec Radford, Jeff Wu, Rewon Child, David Luan, Dario Amodei, and Ilya Sutskever.
\newblock Language models are unsupervised multitask learners.
\newblock 2019.

\bibitem[Rajani et~al.(2020)Rajani, Krause, Yin, Niu, Socher, and Xiong]{rajani2020explaining}
Nazneen~Fatema Rajani, Ben Krause, Wengpeng Yin, Tong Niu, Richard Socher, and Caiming Xiong.
\newblock Explaining and improving model behavior with $k$ nearest neighbor representations, 2020.

\bibitem[Robbins and Monro(1951)]{robbins1951stochastic}
Herbert Robbins and Sutton Monro.
\newblock A stochastic approximation method.
\newblock \emph{The annals of mathematical statistics}, pages 400--407, 1951.

\bibitem[Robertson et~al.(1995)Robertson, Walker, Jones, Hancock-Beaulieu, and Gatford]{robertson1995okapi}
Stephen~E Robertson, Steve Walker, Susan Jones, Micheline~M Hancock-Beaulieu, and Mike Gatford.
\newblock {Okapi at TREC-3}.
\newblock \emph{Nist Special Publication Sp}, 109:\penalty0 109, 1995.

\bibitem[Schioppa et~al.(2022)Schioppa, Zablotskaia, Vilar, and Sokolov]{schioppa2022scaling}
Andrea Schioppa, Polina Zablotskaia, David Vilar, and Artem Sokolov.
\newblock Scaling up influence functions.
\newblock In \emph{Proceedings of the AAAI Conference on Artificial Intelligence}, volume~36, pages 8179--8186, 2022.

\bibitem[Schioppa et~al.(2024)Schioppa, Filippova, Titov, and Zablotskaia]{schioppa2023theoretical}
Andrea Schioppa, Katja Filippova, Ivan Titov, and Polina Zablotskaia.
\newblock Theoretical and practical perspectives on what influence functions do.
\newblock \emph{Advances in Neural Information Processing Systems}, 36, 2024.

\bibitem[Shapley(1953)]{shapley1953value}
Lloyd Shapley.
\newblock A value for $n$-person games.
\newblock 1953.

\bibitem[Singla et~al.(2023)Singla, Sandoval-Segura, Goldblum, Geiping, and Goldstein]{singla2023simple}
Vasu Singla, Pedro Sandoval-Segura, Micah Goldblum, Jonas Geiping, and Tom Goldstein.
\newblock A simple and efficient baseline for data attribution on images.
\newblock \emph{arXiv preprint arXiv:2311.03386}, 2023.

\bibitem[Smith et~al.(1988)Smith, Everhart, Dickson, Knowler, and Johannes]{smith1988using}
Jack~W Smith, James~E Everhart, WC~Dickson, William~C Knowler, and Robert~Scott Johannes.
\newblock Using the adap learning algorithm to forecast the onset of diabetes mellitus.
\newblock In \emph{Proceedings of the annual symposium on computer application in medical care}, page 261. American Medical Informatics Association, 1988.

\bibitem[Spearman(1987)]{spearman1987proof}
Charles Spearman.
\newblock The proof and measurement of association between two things.
\newblock \emph{The American journal of psychology}, 100\penalty0 (3/4):\penalty0 441--471, 1987.

\bibitem[Tieleman and Hinton(2012)]{RMSProp}
Tijmen Tieleman and Geoffrey Hinton.
\newblock Lecture 6.5-rmsprop: Divide the gradient by a running average of its recent magnitude.
\newblock \emph{COURSERA: Neural networks for machine learning}, 4\penalty0 (2):\penalty0 26--31, 2012.

\bibitem[Tsanas and Little(2009)]{misc_parkinsons_telemonitoring_189}
Athanasios Tsanas and Max Little.
\newblock {Parkinsons Telemonitoring}.
\newblock UCI Machine Learning Repository, 2009.
\newblock {DOI}: https://doi.org/10.24432/C5ZS3N.

\bibitem[Vaswani et~al.(2017)Vaswani, Shazeer, Parmar, Uszkoreit, Jones, Gomez, Kaiser, and Polosukhin]{vaswani2017attention}
Ashish Vaswani, Noam Shazeer, Niki Parmar, Jakob Uszkoreit, Llion Jones, Aidan~N Gomez, {\L}ukasz Kaiser, and Illia Polosukhin.
\newblock Attention is all you need.
\newblock \emph{Advances in neural information processing systems}, 30, 2017.

\bibitem[Wang et~al.(2019)Wang, Singh, Michael, Hill, Levy, and Bowman]{wang2018glue}
Alex Wang, Amanpreet Singh, Julian Michael, Felix Hill, Omer Levy, and Samuel~R. Bowman.
\newblock {GLUE}: {A} multi-task benchmark and analysis platform for natural language understanding, 2019.

\bibitem[Wang and Jia(2023)]{wang2023data}
Jiachen~T Wang and Ruoxi Jia.
\newblock {Data Banzhaf}: {A} robust data valuation framework for machine learning.
\newblock In \emph{International Conference on Artificial Intelligence and Statistics}, pages 6388--6421. PMLR, 2023.

\bibitem[Wang et~al.(2024)Wang, Zhu, Wang, Jia, and Mittal]{wang2024privacy}
Jiachen~Tianhao Wang, Yuqing Zhu, Yu-Xiang Wang, Ruoxi Jia, and Prateek Mittal.
\newblock A privacy-friendly approach to data valuation.
\newblock \emph{Advances in Neural Information Processing Systems}, 36, 2024.

\bibitem[Weisberg and Cook(1982)]{weisberg1982residuals}
Sanford Weisberg and R~Dennis Cook.
\newblock Residuals and influence in regression.
\newblock 1982.

\bibitem[Wolf et~al.(2020)Wolf, Debut, Sanh, Chaumond, Delangue, Moi, Cistac, Rault, Louf, Funtowicz, Davison, Shleifer, von Platen, Ma, Jernite, Plu, Xu, Scao, Gugger, Drame, Lhoest, and Rush]{wolf-etal-2020-transformers}
Thomas Wolf, Lysandre Debut, Victor Sanh, Julien Chaumond, Clement Delangue, Anthony Moi, Pierric Cistac, Tim Rault, Rémi Louf, Morgan Funtowicz, Joe Davison, Sam Shleifer, Patrick von Platen, Clara Ma, Yacine Jernite, Julien Plu, Canwen Xu, Teven~Le Scao, Sylvain Gugger, Mariama Drame, Quentin Lhoest, and Alexander~M. Rush.
\newblock Transformers: State-of-the-art natural language processing.
\newblock In \emph{Proceedings of the 2020 Conference on Empirical Methods in Natural Language Processing: System Demonstrations}, pages 38--45, Online, October 2020. Association for Computational Linguistics.
\newblock URL \url{https://www.aclweb.org/anthology/2020.emnlp-demos.6}.

\bibitem[Xia et~al.(2024)Xia, Malladi, Gururangan, Arora, and Chen]{xia2024less}
Mengzhou Xia, Sadhika Malladi, Suchin Gururangan, Sanjeev Arora, and Danqi Chen.
\newblock {LESS}: Selecting influential data for targeted instruction tuning, 2024.

\bibitem[Xiao et~al.(2017)Xiao, Rasul, and Vollgraf]{xiao2017fashion}
Han Xiao, Kashif Rasul, and Roland Vollgraf.
\newblock {Fashion-MNIST}: {A} novel image dataset for benchmarking machine learning algorithms, 2017.

\bibitem[Yeh et~al.(2018)Yeh, Kim, Yen, and Ravikumar]{yeh2018representer}
Chih-Kuan Yeh, Joon Kim, Ian En-Hsu Yen, and Pradeep~K Ravikumar.
\newblock Representer point selection for explaining deep neural networks.
\newblock \emph{Advances in neural information processing systems}, 31, 2018.

\bibitem[Yeh et~al.(2022)Yeh, Taly, Sundararajan, Liu, and Ravikumar]{yeh2022first}
Chih-Kuan Yeh, Ankur Taly, Mukund Sundararajan, Frederick Liu, and Pradeep Ravikumar.
\newblock First is better than last for language data influence.
\newblock \emph{Advances in Neural Information Processing Systems}, 35:\penalty0 32285--32298, 2022.

\bibitem[Yeh(2007)]{misc_concrete_compressive_strength_165}
I-Cheng Yeh.
\newblock {Concrete Compressive Strength}.
\newblock {UCI Machine Learning Repository}, 2007.
\newblock {DOI}: https://doi.org/10.24432/C5PK67.

\bibitem[Zagoruyko and Komodakis(2017)]{zagoruyko2016wide}
Sergey Zagoruyko and Nikos Komodakis.
\newblock Wide residual networks, 2017.

\bibitem[Zhang et~al.(2023)Zhang, Ippolito, Lee, Jagielski, Tram{\`e}r, and Carlini]{zhang2021counterfactual}
Chiyuan Zhang, Daphne Ippolito, Katherine Lee, Matthew Jagielski, Florian Tram{\`e}r, and Nicholas Carlini.
\newblock Counterfactual memorization in neural language models.
\newblock \emph{Advances in Neural Information Processing Systems}, 36:\penalty0 39321--39362, 2023.

\bibitem[Zheng et~al.(2023)Zheng, Pang, Du, Jiang, and Lin]{zheng2023intriguing}
Xiaosen Zheng, Tianyu Pang, Chao Du, Jing Jiang, and Min Lin.
\newblock Intriguing properties of data attribution on diffusion models.
\newblock In \emph{International Conference on Learning Representations}, 2023.

\end{thebibliography}

\newpage
\appendix

\begin{appendices}

\section{Limitations of Leave-One-Out Estimates}
\label{app:loo}

The computation of leave-one-out (\loo) scores in \Cref{eq:loo} presents several computational and conceptual challenges for neural networks. Firstly, calculating the \loos score for all training data points requires retraining the model $N$ times, where $N$ is the size of the training dataset. This process can be prohibitively expensive for large datasets and network architectures.

Moreover, the formulation of \loos assumes that an optimal solution to \Cref{eq:erm} exists, is unique, and can be precisely computed, and that TDA is performed on this optimal solution.  However, within the context of neural networks, these assumptions often do not hold, leading to ambiguities in the computation of \loos estimates. Previous works have investigated various \loos variants as a means to establish counterfactual ground truths \citep{koh2017understanding,basu2020influence,sogaard2021revisiting,jia2021scalability,bae2022if,epifano2023revisiting,nguyen2023bayesian}. For example, \citet{koh2017understanding} and \citet{basu2020influence} considered formulating the \loos ground truth by training the network for an additional number of steps from the final parameters $\finalParams$ without a specific training data point. However, as noted by \citet{bae2022if}, these estimates may reflect the effect of training the network for additional steps instead of model retraining without a data point, especially when the network has not converged.

A more standardized extension of \loos for neural networks is the \emph{expected leave-one-out} (\textsc{ELOO}) retraining \citep{sogaard2021revisiting}, formulated as:
\begin{align}
    \attrib_{\textsc{ELOO}} (\dataPoint_q, \dataPoint_m, \trainingData; \hyperParams) \coloneq \E_\randomness \left[ f(\dataPoint_q, \finalParams(\trainingData \setminus \{ \dataPoint_m \}; \hyperParams, \randomness)) \right] - \E_\randomness \left[ f(\dataPoint_q, \finalParams(\trainingData; \hyperParams, \randomness)) \right],\label{eq:eloo}
\end{align}
where $\hyperParams$ denotes the hyperparameters used to train the model, and the expectation is taken over the randomness in the training process (typically estimated by retraining the network $R$ times). Note that the \textsc{ELOO} can also be seen as the ground truth for the linear datamodeling score (LDS) (defined in \Cref{subsec:tda_evaluate}) with $\alpha = 1 - \sfrac{1}{N}$. Past works have shown the unreliability of \textsc{ELOO} estimates due to the stochasticity in model retraining (\eg, model initialization and batch ordering) \citep{sogaard2021revisiting,epifano2023revisiting,nguyen2023bayesian}. Specifically, \citet{nguyen2023bayesian} observed that the noise from retraining often overshadows the actual signal of removing a single data point, as the effect of removing a single training data point typically has a minor impact on the trained model. In \Cref{subsec:lds_eval}, we also observe that the LDS significantly drops at $\alpha = 1 - \sfrac{1}{N}$, suggesting that the counterfactual ground truth for removing a single data point can be difficult to obtain.
\section{Experimental Setup}
\label{app:experiment_details}

This section describes the experimental setup used to obtain the results presented in \Cref{sec:experiments}. This includes a description of each task (\Cref{app:experiment_details_datasets_models}) and the methodology for computing the linear datamodeling score (LDS) (\Cref{app:lds}). Implementation details of the subset removal counterfactual evaluation and baseline techniques are provided in \Cref{app:counter} and \Cref{app:baseline}, respectively. All experiments were conducted using \textsc{PyTorch} version $2.1.0$ \citep{paszke2017automatic}.

\subsection{Datasets and Models}
\label{app:experiment_details_datasets_models}

We conducted systematic hyperparameter optimization for all tasks. This process involved conducting grid searches to find hyperparameter configurations that achieve the best average validation performance (accuracy for classification tasks and loss for others). The average validation performance was obtained by retraining the network $5$ times using different random seeds. For models trained with SGD with a heavy ball momentum of $0.9$ (SGDm), our search spaces for learning rate and weight decay were \{$3\text{e-}1$, $1\text{e-}1$, $3\text{e-}2$, $1\text{e-}2$, $3\text{e-}3$, $1\text{e-}3$, $3\text{e-}4$, $1\text{e-}4$, $3\text{e-}5$, $1\text{e-}5$\} and $\{3\text{e-}2$, $1\text{e-}2$, $3\text{e-}3$, $1\text{e-}3$, $3\text{e-}4$, $1\text{e-}4$, $3\text{e-}5$, $1\text{e-}5$, $0.0$\}, respectively. For models trained with AdamW \citep{Loshchilov2017decoupled}, the search spaces were \{$1\text{e-}2$, $3\text{e-}3$, $1\text{e-}3$, $3\text{e-}4$, $1\text{e-}4$, $3\text{e-}5$, $1\text{e-}5$\} for learning rate and \{$3\text{e-}2$, $1\text{e-}2$, $3\text{e-}3$, $1\text{e-}3$, $3\text{e-}4$, $1\text{e-}4$, $3\text{e-}5$, $1\text{e-}5$, 0.0\} for weight decay. In cases where the original experimental setup from which we adapted had a pre-specified learning rate and weight decay, these hyperparameters were incorporated into our search space.

\paragraph{UCI Datasets (Regression).} For regression tasks, we used the Concrete \citep{misc_concrete_compressive_strength_165} and Parkinson \citep{misc_parkinsons_telemonitoring_189} datasets from the UCI Machine Learning Repository \citep{dua2019uci}. Both datasets were pre-processed to have a zero mean and unit variance for input features and targets. We trained a three-layer multilayer perceptron (MLP), where each layer consisted of $128$ hidden units and the \textsc{ReLU} activation function. The models were optimized using SGDm for $20$ epochs with a batch size of $32$ and a constant learning rate schedule. A learning rate of $3\text{e-}2$ and a weight decay of $1\text{e-}5$ were used for the Concrete dataset. For the Parkinson dataset, the learning rate was set to $1\text{e-}2$ with a weight decay value of $3\text{e-}5$. We saved $6$ intermediate checkpoints throughout training. For the noisy Concrete (Concrete-N) dataset, we randomly modified $30\%$ of the targets by sampling from a Normal distribution with zero mean and unit variance. We used the same hyperparameters but trained the models for $3$ epochs.

\paragraph{MNIST \& FashionMNIST (Image Classification).} Following the experimental setup from \citet{koh2017understanding} and \citet{bae2022if}, we trained a three-layer multilayer perceptron (MLP) on approximately $10\%$ of MNIST \citep{lecun2010mnist} and FashionMNIST \citep{xiao2017fashion} datasets. Smaller versions of these datasets were used to compute the counterfactual ground truth more efficiently. The models were trained with SGDm for $20$ epochs with a batch size of $64$ and a constant learning rate. The learning rate and weight decay were set for both datasets to $3\text{e-}2$ and $1\text{e-}3$, respectively. We saved $6$ checkpoints during training and utilized them for \textsc{TracIn} and \textsc{Source}. For the noisy FashionMNIST (FashionMNST-N) experiment in \Cref{subsec:failure_exp}, we randomly relabeled $30\%$ of the training dataset. The network was only trained for $3$ epochs with a learning rate $1\text{e-}2$ and weight decay $3\text{e-}5$. 

\paragraph{CIFAR-10 (Image Classification).} For the CIFAR-10 dataset \citep{krizhevsky2009learning}, we trained the ResNet-9 model \citep{he2016deep},\footnote{\url{https://github.com/MadryLab/trak/blob/main/examples/cifar_quickstart.ipynb}.} following the standard data augmentation procedure from \citet{zagoruyko2016wide}. This included extracting images from a random $32 \times 32$ crop after applying zero-padding of $4$ pixels, with a $50\%$ probability of horizontal flipping. The network was trained for $25$ epochs using SGDm with a batch size of $512$ and a cyclic learning rate schedule, peaking at $0.5$. The initial learning rate was set to $0.4$ with a weight decay of $1\text{e-}3$, and $6$ intermediate checkpoints were saved throughout training. 

\paragraph{GLUE (Text Classification).} We fine-tuned the BERT model \citep{devlin2018bert} on SST-2, RTE, and QNLI datasets from the GLUE benchmark \citep{wang2018glue} with the training script from the \texttt{Transformers} library \citep{wolf-etal-2020-transformers}.\footnote{\url{https://github.com/huggingface/transformers/blob/main/examples/pytorch/text-classification/run_glue_no_trainer.py}.} Following the experimental setup from \citet{park2023trak}, we capped the training dataset at a maximum of $51200$ examples to compute the LDS efficiently. However, we did not modify the original architecture (\eg,\ removing the last \textsc{Tanh} layer) and trained the network with the AdamW optimizer. The weight decay was set to $1\text{e-}2$ for all tasks, and the learning rates were set as follows: $3\text{e-}5$ for SST-2, $1\text{e-}5$ for QNLI, and $2\text{e-}5$ for RTE. We saved $6$ intermediate checkpoints for each training run.

\paragraph{WikiText-2 (Language Modeling).} For the language modeling task, we fine-tuned the GPT-2 model \citep{radford2019language} using the WikiText-2 dataset \citep{merity2016pointer}. We followed the training script from the Transformer library but set the maximum sequence length to $512$.\footnote{\url{https://github.com/huggingface/transformers/blob/main/examples/pytorch/language-modeling/run_clm_no_trainer.py}.} During fine-tuning with AdamW, we saved $6$ intermediate checkpoints for data attribution. The learning rate, weight decay, and batch size were set to $3\text{e-}5$, $1\text{e-}2$, and $8$, respectively.

\paragraph{RotatedMNIST \& PACS (Image Classification).} We used the RotatedMNIST dataset \citep{ghifary2015domain} and the PACS dataset \citep{li2017deeper}, following the data pre-processing procedures from \citet{gulrajani2020search}.\footnote{\url{https://github.com/facebookresearch/DomainBed}.} The training process was divided into two distinct stages for both tasks. During the initial stage of the training, we trained the network with the dataset $\trainingData_1$, while the second stage used dataset $\trainingData_2$. For RotatedMNIST, the first dataset $\trainingData_1$ was comprised of images rotated at $0$, $15$, $45$, and $60$ degrees, whereas the second dataset $\trainingData_2$ contained images rotated at $30$ degrees. We trained a three-layer MLP for $30$ ($20$/$10$) epochs using SGDm and a batch size of $128$. The learning rate and weight decay were set to $1\text{e-}1$ and $1\text{e-}5$. For PACS, the first dataset $\trainingData_1$ included images from the cartoon, photo, and sketch categories, and the second dataset $\trainingData_2$ had art paintings. We fine-tuned ResNet-50 \citep{he2016deep}, initialized from the pre-trained parameters,\footnote{\url{https://pytorch.org/vision/main/models/generated/torchvision.models.resnet50.html}.} using SGDm for $40$ ($30$/$10$) epochs with a batch size of $128$, a learning rate of $1\text{e-}4$, and a weight decay of $3\text{e-}5$. 

\subsection{Linear Datamodeling Score}
\label{app:lds}

We follow a methodology proposed by \citet{park2023trak} to compute the linear datamodeling score (LDS). Let $\hyper$ represent the set of hyperparameters used for training the model on a specified task, such as the choice of optimizer and the number of training epochs. Let $\alpha \in (0, 1)$ denote the data sampling ratio. The process for obtaining the LDS involves several steps:
\begin{enumerate}
    \item We generate $M$ data subsets, denoted as $\{\mathcal{S}_j\}_{j=1}^M$, each being a uniformly sampled subset of the original training dataset $\trainingData$. Each subset $\mathcal{S}_j \subset \trainingData$ contains $\lceil \alpha N \rceil$ data points, where $N$ denotes the total number of training data points.
    \item For each data subset $\mathcal{S}_j$, the model is trained $R$ times using different random seeds $\{\randomness_r\}_{r=1}^R$ (\eg,\ model initialization and batch ordering).
    \item Given an attribution method $\attrib$ and a query example $\dataPoint_q$, we measure the Spearman correlations \citep{spearman1987proof} between the prediction and the estimated expected measurable quantity:
    \begin{equation}    
        \begin{aligned}
            \boldsymbol{\rho}  \left( \left\{ \frac{1}{R} \sum_{r=1}^{R} f(\dataPoint_q, \finalParams (\trainingDataSubset_j; \hyper, \randomness_r)) : j \in [M] \right\}, \left\{g_\tau (\dataPoint_q, \trainingDataSubset_j, \trainingData; \hyperParams): j \in [M] \right\} \right),
    \end{aligned}
    \end{equation}
    where $g$ represents the group attribution prediction, expressed as:
    \begin{align}
        g_\tau (\dataPoint_q, \trainingDataSubset, \trainingData; \hyperParams) \coloneqq \sum_{ \dataPoint \in \trainingDataSubset} \attrib (\dataPoint_q, \dataPoint, \trainingData; \hyperParams).
    \end{align}
    \item To obtain the final LDS, we average the correlations over a set of query data points (up to $2000$ in our experiments) and report the score with 95\% bootstrap confidence intervals, which accounts for resampling of the data subset $\trainingDataSubset_j$.
\end{enumerate}

For a given data sampling ratio $\alpha$, the networks must be retrained $MR$ times in total to compute the LDS ground truth. In our experiments, we used 100 subsets ($M = 100$). The repeat $R$ was set to $100$ for UCI regression tasks, $10$ for MNIST classification tasks, $20$ for CIFAR-10 image classification task, $5$ for GLUE text classification and WikiText language modeling task, and $20$ for RotatedMNIST and PACS image classification tasks. We used the largest feasible $R$ based on our computational budget because we observed improvements in LDS for baseline techniques (especially \textsc{Trak}, \textsc{IF}, and \textsc{Source}) with larger $R$.

\subsection{Subset Removal Counterfactual Evaluation}
\label{app:counter}

For the subset removal counterfactual evaluation, we first train the model with the full dataset $\trainingData$ under different random choices (over $5$ random seeds) and select $100$ test data points correctly classified on all random choices. Then, for each test data point and attribution technique, we remove the top-$k$ data points from the pre-defined interval $k_1, \dots, k_I$ (such that $k_1 < \dots < k_I$), as indicated as highly positively influential by the data attribution technique, retrain the network with this modified dataset, and examine if the original test data point gets misclassified on average under different random choices (over $3$ random seeds). Finally, for each value of $k$ in the pre-defined interval, we report the fraction of test data points that get misclassified after removing at most top-$k$ training data points and retraining the network with the modified dataset.

For each TDA technique, this process requires retraining the model $100 \times I \times 3$ times, where $I$ is the pre-defined interval size. We set $I = 6$ for all experiments, leading to the retraining of the model $1800$ times. To reduce the computational cost, we start from the smallest subset removal size $k_1$, and if the test data point gets misclassified under the current subset, we do not consider it for the larger subset removal size (\eg,\ $k_2$). Hence, this can be seen as the fraction of test data points that get misclassified by removing at most $k$ training data points (evaluated at a fixed interval). We note that \citet{singla2023simple} instead use a bisection search to find the smallest subset size in which a test data point can be misclassified, whereas \citet{ilyas2022datamodels} use more fine-grained intervals with more number of seeds (\eg, $8$ intervals and $20$ seeds). We used \methods and baseline techniques described in \Cref{app:baseline} to identify positively influential training data points. We also included a \textsc{Random} baseline, where we removed the training data points belonging to the same class as the target test example.

\subsection{Baselines}
\label{app:baseline}

This section describes the baseline techniques used in \Cref{sec:experiments}. Unless specified otherwise, we describe them in the context of a single-training-run estimator, where the TDA techniques use the final parameters $\finalParams$ obtained with hyperparameters $\hyperParams$ and some random choice $\randomness$ (the multiple-training-runs estimators simply average the TDA scores obtained from models trained with different random choices $\randomness$).

\paragraph{Representation Similarity (\textsc{RepSim}).} Representation similarity technique \citep{caruana1999case} evaluates the importance of a training data point $\dataPoint_m \in \trainingData$ to a specific query data point $\dataPoint_q$ by comparing the latent representations of these data point pairs. This can be formulated as follows:
\begin{align}
    \tau_{\textsc{RepSim}} (\dataPoint_q, \dataPoint_m, \trainingData; \hyperParams) \coloneqq \text{similarity}(\phi_{\finalParams} (\dataPoint_q), \phi_{\finalParams} (\dataPoint_m)).
\end{align}
Here, $\text{similarity}(\mathbf{v}_1, \mathbf{v}_2)$, where $\mathbf{v}_1$ and $\mathbf{v}_2$ are some vectors, is typically defined through the $\ell_2$ metric, dot metric, or cosine metric \citep{hanawa2020evaluation}. In our experiments, the function $\phi_{\finalParams}(\dataPoint)$ was designed to map a data point to its last hidden activations (before the final output layer), using a forward pass through the final parameters $\finalParams$. We used the cosine metric to compute the attribution score but observed similar performance when using the $\ell_2$ metric, aligning with observations in previous studies \citep{ilyas2022datamodels,park2023trak,singla2023simple}.

\paragraph{\textsc{TracIn}.} We used the \textsc{TracInCP} estimator from \citet{pruthi2020estimating}, defined as:
\begin{align}
    \tau_{\textsc{TracIn}} (\dataPoint_q, \dataPoint_m, \trainingData; \hyperParams) \coloneqq \sum_{k=1}^C \LR_k \cdot \nabla_{\params} f(\dataPoint_q, \paramsCheckpoint_k) \cdot \nabla_{\params} \loss (\dataPoint_m, \paramsCheckpoint_k),
    \label{eq:tracin}
\end{align}
where $C$ represents the number of checkpoints, $\paramsCheckpoint_k$ represents the parameters at the $k$-th checkpoint, and $\LR_k$ is the learning rate applied at the corresponding checkpoint. The last checkpoint is typically set to the final model parameters $\finalParams$. While there is an option to compress the gradients using a random projection as suggested by \citet{pruthi2020estimating}, our experiments used the full gradients to obtain a stronger baseline. The checkpoint selection details are described in \Cref{app:experiment_details_datasets_models}.

\paragraph{Influence Functions (\textsc{IF}).} As detailed in \Cref{subsec:influence_functions}, training data attribution with influence functions is formulated as follows:
\begin{align}
    \tau_{\textsc{IF}} (\dataPoint_q, \dataPoint_m, \mathcal{D}; \hyperParams) \coloneqq \nabla_{\params} f(\dataPoint_q, \finalParams)^\top \hessian^{-1} \nabla_{\params} \loss(\dataPoint_m, \finalParams),
\end{align}
where $\hessian$ denotes the Hessian of the cost at the final parameters $\finalParams$. To make influence functions scalable to large neural networks, we used the Eigenvalue-corrected Kronecker-Factored Approximate Curvature (EK-FAC) parameterization \citep{george2018fast} to approximate the Hessian, as proposed by \citet{grosse2023studying}. We refer readers to \citet{grosse2023studying} and \Cref{app:implementation_details} for details on the EK-FAC computation. Relatedly, \citet{schioppa2022scaling} use Arnoldi iterations \citep{arnoldi1951principle}, and \citet{kwon2023datainf} utilize the parameter-efficient fine-tuning (PEFT) \citep{hu2021lora} strategy to efficiently approximate influence functions.

While \citet{grosse2023studying} only consider the computation of influence scores to the MLP layers of transformers \citep{vaswani2017attention}, in our experiments, we extended this computation to include the attention layers as well. We excluded layer normalization, batch normalization, and embedding layers from the influence computation. Influence functions have an additional hyperparameter $\lambda > 0$, which is used to compute the damped inverse Hessian-vector product (IHVP), denoted as $(\hessian + \lambda \eye)^{-1} \mathbf{v}$ for some vector $\mathbf{v}$. We used a small damping term for consistency with \textsc{Trak} \citep{park2023trak} and set it to $1\text{e-}8$ to avoid numerical instability (note that \textsc{Trak} sets the damping term to $0$). All experiments were conducted based on the \texttt{Kronfluence} repository.\footnote{\url{https://github.com/pomonam/kronfluence}.}

\paragraph{\textsc{Trak}.} In contrast to the traditional formulation of influence functions, \textsc{Trak} \citep{park2023trak} leverages random projections \citep{johnson1986extensions}, Generalized Gauss-Newton approximation, and ensembling for data attribution. Specifically, given a random projection matrix $\mathbf{P} \sim \mathcal{N}(0, 1)^{M \times K}$, where $K$ denotes the projection dimension, the final model parameters $\finalParams$, and a model output function $f(\dataPoint, \params)$, \textsc{Trak} projects all training and query gradients into $K$-dimensional vectors. The feature map is defined as:
\begin{align}
    \phi(\dataPoint) \coloneqq \mathbf{P}^\top \nabla_{\params} f(\dataPoint, \finalParams).
\end{align}
We further define $\boldsymbol{\Phi} \coloneqq [\phi_1; \dots; \phi_N] \in \mathbb{R}^{N \times K}$ as stacked projected gradients for all training data points, where each $\phi_i$ corresponds to $\phi(\dataPoint_i)$. Subsequently, \textsc{Trak}'s single model estimator is formulated as:
\begin{align}
    \tau_{\textsc{Trak}} (\dataPoint_q, \cdot, \trainingData; \hyperParams) \coloneqq \phi(\dataPoint_q)^\top (\boldsymbol{\Phi}^\top \boldsymbol{\Phi})^{-1} \boldsymbol{\Phi}^\top \mathbf{Q},
\end{align}
with $\mathbf{Q}$ being a $N \times N$ diagonal matrix for weightings. Here, $\tau_{\textsc{Trak}}$ represents a vector of dimension $N$, containing attribution score for each training data point. \textsc{Trak} uses an ensemble of single model estimators, each derived from models trained with distinct configurations and projection matrices. We refer readers to \citet{park2023trak} and \citet{engstrom2024dsdm} for detailed derivations and discussions of \textsc{Trak}.

We used the final checkpoints for \textsc{Trak} in our experimental setup involving a single model. We computed \textsc{Trak} using the last checkpoint of $10$ differently trained models (each trained with $50\%$ of the dataset) for experiments with multiple model setups. \textsc{Trak} has a hyperparameter that determines the dimension of the random projection $K$. We set the projection dimension to $20480$ for ResNet-9 and RotatedMNIST, $8192$ for ResNet-50 on the PACS dataset, $1024$ for BERT trained on the RTE dataset and $512$ for MLP trained on the Concrete dataset (due to the datasets' smaller size), and $4096$ for all other tasks. All experiments were conducted using \textsc{Trak}'s official implementation.\footnote{\url{https://github.com/MadryLab/trak}.}

\paragraph{Empirical Influence (\textsc{EI}).} To compute the empirical influence (\textsc{Downsampling}) \citep{feldman2020neural}, we first create $M$ data subsets $\{ \mathcal{S}_j \}_{j=1}^M$, each being a uniformly sampled subset of the original training dataset. Each subset $\mathcal{S}_i$ contains $\lceil \alpha N \rceil$ data points, where $\alpha \in (0, 1)$ is the data sampling ratio. Given a training data point $\dataPoint_m \in \trainingData$, we define $M_m$ as the total number of data subsets containing $\dataPoint_m$. The empirical influence scores are formulated as follows:
\begin{align}
    \tau_{\textsc{EI}} (\dataPoint_q, \dataPoint_m, \mathcal{D}; \hyperParams) \coloneqq &\frac{1}{M - M_m} \sum_{j=1}^M \mathbb{1}[\dataPoint_m \notin \trainingDataSubset_j] f(\dataPoint_q, \finalParams(\trainingDataSubset_j; \hyperParams, \randomness_j)) \label{eq:ei-r}\\
    &\quad - \frac{1}{M_m} \sum_{j=1}^M \mathbb{1}[\dataPoint_m \in \trainingDataSubset_j] f(\dataPoint_q, \finalParams(\trainingDataSubset_j; \hyperParams, \randomness_j)), \label{eq:ei-k}
\end{align}
where $\mathbb{1}[\cdot]$ is an indicator function to determine if the training data point $\dataPoint_m$ is contained in the $j$-th data subset $\trainingDataSubset_j$. Intuitively, \Cref{eq:ei-r} computes the averaged query measurement when data point $\dataPoint_m$ is not used in training, whereas \Cref{eq:ei-k} computes the averaged measurement when the data point is used in training. Following \citet{zheng2023intriguing}, we created $512$ data subsets ($M = 512$) with a sampling ratio $\alpha = 0.5$, which requires retraining the model $512$ times with $50\%$ of training data points removed.

\paragraph{\textsc{Hydra}.} We used the fast version of \textsc{Hydra} \citep{chen2021hydra}, formulated as:
\begin{align}
    \tau_{\textsc{Hydra}} (\dataPoint_q, \dataPoint_m, \trainingData; \hyperParams) \coloneqq \sum_{k=0}^{T-1} \LR_k \cdot \mathbb{1}[\dataPoint_m \in \mathcal{B}_k] \cdot \nabla_{\params} f(\dataPoint_q, \finalParams) \cdot \nabla_{\params} \loss (\dataPoint_m, \params_k),
    \label{eq:hydra}
\end{align}
where $T$ represents the total number of gradient update steps, $\params_k$ denotes the parameters at the $k$-th iteration, and $\eta_k$ is the corresponding learning rate. Here, $\mathcal{B}_k$ denotes the batch of data points used at the corresponding update, and $\mathbb{1}[\dataPoint_m \in \mathcal{B}_k]$ is the indicator function for having selected $\dataPoint_m$ in the update. Note that \textsc{Hydra} requires storing all parameter vectors used for training. We refer readers to \citet{hammoudeh2022training} for derivations and detailed discussions of \textsc{Hydra}.

\section{\textsc{Source} with Preconditioning Matrix}
\label{app:general}

In \Cref{subsec:unroll}, we motivated our proposed algorithm, \method, for cases where the parameters are optimized using stochastic gradient descent (SGD). In this section, we present the formulation of \methods when preconditioned optimizers, such as RMSProp \citep{RMSProp}, Adam \citep{kingma2014adam}, and K-FAC \citep{martens2015optimizing}, are used to train the model.

To investigate the impact of removing a training data point $\dataPoint_m \in \trainingData$, we follow a similar derivation as in \Cref{subsec:unroll}, but now considering the preconditioning matrix:
\begin{align}
    \params_{k+1} (\epsilon) \leftarrow \params_k (\epsilon) - \frac{\LR_k}{B} \mathbf{P}_k \left( \sum_{i=1}^B  (1 + \delta_{ki} \epsilon) \nabla_{\params} \loss(\dataPoint_{ki}, \params_k (\epsilon))\right),
\end{align}
where $\mathbf{P}_k$ is a (positive definite) preconditioning matrix and $\delta_{ki} \coloneq \mathbb{1}[\dataPoint_{ki} = \dataPoint_m]$ is the indicator function for having selected $\dataPoint_m$. 

By applying the chain rule of derivatives, the contribution of iteration $k$ to the total derivative can be found by multiplying all the Jacobian matrices along the backward accumulation path, giving the value $-\tfrac{\LR_k}{\numBatch} \jacobian_{k+1:T} \precond_k \grad_k$. Hence, by applying the linearity of expectation, the expected total derivative of the terminal parameters $\params_T$ with respect to the perturbation $\epsilon$ is expressed as:
\begin{equation}
\begin{aligned}
    \mathbb{E} \left[ \frac{\mathrm{d} \params_{T}}{\mathrm{d} \epsilon} \right] = -\sum_{k=0}^{T-1} \frac{\eta_k}{B} \mathbb{E}[\delta_k \jacobian_{k+1:T} \mathbf{P}_k \grad_k],
\end{aligned}
\end{equation}
where we have:
\begin{equation}
  \begin{gathered}
    \jacobian_k \coloneq \frac{\mathrm{d} \params_{k+1}}{\mathrm{d} \params_k} = \eye - \LR_k \precond_k \hessian_k \\
    \jacobian_{k:k'} \coloneq \frac{\mathrm{d} \params_{k'}}{\mathrm{d} \params_k} = \jacobian_{k'-1} \cdots \jacobian_{k+1} \jacobian_{k}\\
    \grad_k \coloneq \nabla_{\params} \mathcal{L}(\dataPoint_m, \params_k).
  \end{gathered}
\end{equation}

As discussed in \Cref{subsec:segment}, we group the training trajectories into multiple segments to approximate the expected total derivative for each segment with statistical summaries thereof. In addition to the approximations introduced in \Cref{subsec:segment}, we approximate preconditioning matrices as stationary within a segment and represent it as $\precondApprox_\ell \coloneq \precond_k$ for $T_{\ell-1} \leq k < T_{\ell}$. 

\paragraph{Approximation of $\mathbb{E}[\segment_\ell]$.} We approximate $\E[\segmentS_\ell]$ in \Cref{eqn:final} as follows:
\begin{equation}
\begin{aligned}
    \mathbb{E} [\segment_\ell] &= \mathbb{E} [\jacobian_{T_{\ell-1}:T_{\ell}}] \approx \left(\eye - \bar{\LR}_\ell \bar{\mathbf{P}}_\ell \bar{\hessian}_\ell \right)^{K_{\ell}} \approx \exp (-\bar{\LR}_\ell K_{\ell} \bar{\mathbf{P}}_\ell \bar{\hessian}_\ell) \\
    &= \precondApprox_\ell^{1/2} \exp (-\bar{\LR}_\ell K_{\ell} \bar{\mathbf{P}}_\ell^{1/2} \bar{\hessian}_\ell \bar{\mathbf{P}}_\ell^{1/2}) \precondApprox_\ell^{-1/2} \coloneqq \bar{\segment}_\ell.
    \label{eqn:adam}
\end{aligned}
\end{equation}
Note that the last line uses the properties of the matrix exponential.\footnote{For a square matrix $\mathbf{M}$ and a square positive definite matrix $\mathbf{D}$, we have $\exp(\mathbf{M}) = \sum_{k=0}^{\infty} \frac{1}{k!} \mathbf{M}^k = \mathbf{D}^{1/2} \left[ \sum_{k=0}^\infty \frac{1}{k!} \left(\mathbf{D}^{-1/2} \mathbf{M} \mathbf{D}^{1/2} \right)^k \right] \mathbf{D}^{-1/2} = \mathbf{D}^{1/2} \exp(\mathbf{D}^{-1/2} \mathbf{M} \mathbf{D}^{1/2}) \mathbf{D}^{-1/2}$.}

\paragraph{Approximation of $\mathbb{E}[\mathbf{r}_\ell]$.} We further approximate $\mathbb{E}[\mathbf{r}_\ell]$ as follows:
\begin{align}
    \mathbb{E}[\segmentR_\ell] &= \E \left[ \sum_{k=T_{\ell-1}}^{T_\ell - 1} \frac{\eta_k}{B} \delta_k \jacobian_{k+1:T_\ell} \mathbf{P}_k \mathbf{g}_k\right]\\
    &\approx \frac{1}{N} \sum_{k=T_{\ell-1}}^{T_{\ell} - 1} \bar{\LR}_\ell (\eye - \bar{\LR}_\ell \precondApprox_\ell \barHess_\ell)^{T_{\ell} - 1 - k} \precondApprox_\ell \barGrad_\ell \\
    &= \frac{1}{N}(\eye - (\eye - \bar{\LR}_\ell \precondApprox_\ell \barHess_\ell)^{K_{\ell}} )\barHess_\ell^{-1} \barGrad_\ell \\
    &\approx \frac{1}{N}(\eye - \exp(-\bar{\LR}_\ell K_{\ell} \precondApprox_\ell \barHess_\ell) ) \barHess_\ell^{-1} \barGrad_\ell \label{eq:precond_r2}\\
    &= \frac{1}{N} \precondApprox_\ell^{1/2} \underbrace{(\eye - \exp(-\bar{\LR}_\ell K_\ell \mathbf{M}_\ell)) \mathbf{M}_\ell^{-1}}_{\coloneq F_{\mathbf{r}}} \precondApprox_\ell^{1/2} \barGrad_\ell \coloneq \bar{\mathbf{r}}_\ell, \label{eq:precond_r}
\end{align}
where we define $\mathbf{M}_\ell \coloneq \bar{\mathbf{P}}_\ell^{1/2} \bar{\hessian}_\ell \bar{\mathbf{P}}_\ell^{1/2}$ and the last line uses the properties of matrix exponential, as done in \Cref{eqn:adam}. Similarly to our analysis presented in \Cref{subsec:segment}, we can represent $\bar{\mathbf{r}}_\ell$ with the matrix function of $\mathbf{M}_\ell$. Let $\mathbf{M}_\ell = \mathbf{Q} \boldsymbol{\Lambda} \mathbf{Q}^\top$ be the eigendecomposition of $\mathbf{M}_\ell$ and let $\sigma_j$ be the $j$-th eigenvalue of $\mathbf{M}_\ell$. The expression can be seen as applying the matrix function, defined as:
\begin{align}
    F_{\mathbf{r}} (\sigma) \coloneq \frac{1 - \exp{\left(-\bar{\LR}_\ell K_\ell \sigma\right)}}{\sigma}.
\end{align}
The qualitative behavior of $F_{\mathbf{r}}$ can be captured with the function $F_{\rm inv}(\sigma) \coloneq 1/(\sigma + \lambda)$, where $\lambda = \barLR^{-1}_\ell K_{\ell}^{-1}$ (see \Cref{subsec:segment} for details). Hence, one way to understand \Cref{eq:precond_r} is by expressing it as the damped inverse Hessian-vector product (iHVP):
\begin{align}
    \bar{\mathbf{r}}_\ell &\approx \frac{1}{N} \bar{\mathbf{P}}_\ell^{1/2}(\mathbf{M}_\ell + \lambda \eye)^{-1} \bar{\mathbf{P}}_\ell^{1/2} \barGrad_\ell\\
    &= \frac{1}{N} \bar{\mathbf{P}}_\ell^{1/2}(\bar{\mathbf{P}}_\ell^{1/2} \barHess_{\ell} \bar{\mathbf{P}}_\ell^{1/2} + \lambda \eye)^{-1} \bar{\mathbf{P}}_\ell^{1/2} \barGrad_\ell\\
    &= \frac{1}{N} (\barHess_{\ell} + \lambda \precondApprox_\ell^{-1})^{-1} \barGrad_\ell.\label{eq:adam_matrix}
\end{align}
In a case where $\precondApprox_\ell$ is a diagonal matrix, \Cref{eq:adam_matrix} can be seen as a special case for influence functions with a specific diagonal damping term $\lambda \precondApprox_\ell^{-1}$. Using the derived $\bar{\segment}_\ell$ and $\bar{\mathbf{r}}_\ell$, we approximate the total expected derivative using \Cref{eq:unif_segment}.

\section{Implementation Details}
\label{app:implementation_details}

This section describes the Eigenvalue-corrected Kronecker-Factored Approximate Curvature (EK-FAC) \citep{george2018fast} and how we computed \methods using this EK-FAC parameterization. The code for implementing \methods (as well as baseline techniques) will be provided at \url{https://github.com/pomonam/kronfluence}. For details on the EK-FAC approximation specific to influence functions, we refer readers to \citet{grosse2023studying}. 

\subsection{Eigenvalue-corrected Kronecker-Factored Approximate Curvature (EK-FAC)}

Kronecker-Factored Approximate Curvature (K-FAC) \citep{martens2015optimizing} and EK-FAC \citep{george2018fast} introduce a parametric approximation to the Fisher information matrix (FIM) of a neural network, defined as:
\begin{align}
    \mathbf{F} \coloneqq \mathbb{E}_{\boldsymbol{x} \sim p_{\text{data}}, \hat{\boldsymbol{y}} \sim P_{\hat{\boldsymbol{y}} | \boldsymbol{x}} (\params)} \left[ \nabla_{\params} \log p(\hat{\boldsymbol{y}} | \params, \boldsymbol{x}) \nabla_{\params} \log p(\hat{\boldsymbol{y}} | \params, \boldsymbol{x})^\top \right],\label{eq:fim}
\end{align}
where $p_{\text{data}}$ is the data distribution and $P_{\hat{\boldsymbol{y}} | \boldsymbol{x}} (\params)$ is the model's output distribution. For many commonly used loss functions, such as softmax-cross-entropy and squared-error, the FIM is equivalent to the Gauss-Newton Hessian (GNH) \citep{martens2020new}, denoted as $\mathbf{G}$. The GNH can be seen as an approximation to the Hessian $\mathbf{H}$, where the network is linearized around the current parameters \citep{GrosseNNTDChapter4}. Different from the Hessian, the GNH is guaranteed to be positive semi-definite (PSD) when the loss function is convex with respect to the model output.

While K-FAC and EK-FAC were originally formulated for multilayer perceptrons (MLPs), they were later extended to other architectures, such as convolutional neural networks \citep{grosse2016kronecker}, recurrent neural networks \citep{martens2018kronecker}, graph neural networks \citep{izadi2020optimization}, or to be learnable by gradient-based optimizers \citep{bae2022amortized}. We refer readers to \citet{eschenhagen2024kronecker} for a comprehensive overview. This section describes the EK-FAC formulation in the context of MLPs.

Consider a $l$-th layer of the network with input activations $\mathbf{a}_{l-1} \in \mathbb{R}^I$ and pre-activation output $\mathbf{s}_{l} \in \mathbb{R}^O$ such that $\mathbf{s}_l \coloneqq \mathbf{W}_l \mathbf{a}_{l-1}$, where $\mathbf{W} \in \mathbb{R}^{O \times I}$ is the weight matrix (we drop the layer subscript to avoid clutter and ignore the bias term for simplicity). The pseudo-gradient (where the target is sampled from the model's output distribution; see \Cref{eq:fim}) is given by $\mathcal{D} \mathbf{W} \coloneqq \mathcal{D} \mathbf{s} \mathbf{a}^\top$. K-FAC makes two core approximations: (1) layerwise independence approximation, where GNH is approximated as block-diagonal with each block corresponding to GNH of some specific layer, and (2) input activations $\mathbf{a}$ and pseudo-gradient of the pre-activations $\mathcal{D} \mathbf{s}$ are independent under the model's predictive distribution. The layerwise GNH can be approximated as:
\begin{align}
    \mathbf{G} = \mathbb{E} \left[\text{vec}(\mathcal{D} \mathbf{W}) \text{vec}(\mathcal{D} \mathbf{W})^\top \right] = \mathbb{E} \left[\mathbf{a} \mathbf{a}^\top \otimes \mathcal{D}\mathbf{s} \mathcal{D}\mathbf{s}^\top \right] \approx \mathbb{E}[\mathbf{a} \mathbf{a}^\top] \otimes \mathbb{E}\left[\mathcal{D}\mathbf{s} \mathcal{D}\mathbf{s}^\top \right] \coloneqq \mathbf{A} \otimes \mathbf{S}, \label{eq:kfac}
\end{align}
where $\otimes$ denotes the Kronecker product. The matrices $\mathbf{A} \in \mathbb{R}^{I \times I}$ and $\mathbf{S} \in \mathbb{R}^{O \times O}$ in \Cref{eq:kfac} represent the uncentered covariance matrices of the activations and the pseudo-gradients with respect to the pre-activations, respectively. These covariance matrices can be estimated by computing the statistics over many data batches and taking the average.

Denoting the eigendecomposition of these covariance matrices as $\mathbf{A} = \mathbf{Q}_\mathbf{A} \boldsymbol{\Lambda}_\mathbf{A} \mathbf{Q}_\mathbf{A}^\top$ and $\mathbf{S} = \mathbf{Q}_\mathbf{S} \boldsymbol{\Lambda}_\mathbf{S} \mathbf{Q}_\mathbf{S}^\top$, using properties of the Kronecker product, we can express the eigendecomposition of $\mathbf{A} \otimes \mathbf{B}$ as:
\begin{align}
    \mathbf{A} \otimes \mathbf{B} =  (\mathbf{Q}_\mathbf{A} \otimes \mathbf{Q}_\mathbf{S}) (\boldsymbol{\Lambda}_\mathbf{A} \otimes \boldsymbol{\Lambda}_\mathbf{S}) (\mathbf{Q}_\mathbf{A} \otimes \mathbf{Q}_\mathbf{S})^\top. \label{eq:ekfac}
\end{align}

EK-FAC introduces a more accurate approximation to the GNH by introducing a compact representation of the eigenvalues (instead of representing them as the Kronecker product $\boldsymbol{\Lambda}_\mathbf{A} \otimes \boldsymbol{\Lambda}_\mathbf{S}$). The layerwise GNH for EK-FAC is represented as follows:
\begin{align}
    \mathbf{G} \approx (\mathbf{Q}_\mathbf{A} \otimes \mathbf{Q}_\mathbf{S}) \boldsymbol{\Lambda} (\mathbf{Q}_\mathbf{A} \otimes \mathbf{Q}_\mathbf{S})^\top.
\end{align}
Here, the corrected eigenvalues $\boldsymbol{\Lambda} \in \mathbb{R}^{IO \times IO}$ are defined as:
\begin{align}
    \boldsymbol{\Lambda}_{ii} \coloneqq \mathbb{E}[((\mathbf{Q}_{\mathbf{A}} \otimes \mathbf{Q}_{\mathbf{S}}) \text{vec}(\mathcal{D} \mathbf{W}))_i^2].\label{eq:ekfac_eigen}
\end{align}
The corrected eigenvalues in \Cref{eq:ekfac_eigen} minimize the approximation error with the GNH measured by the Frobenius norm, where we refer readers to \citet{george2018fast} for the derivations.

\subsection{EK-FAC Computations for \textsc{SOURCE}}

As detailed in \Cref{subsec:algo}, our practical instantiation of \methods requires averaging the Hessians across checkpoints within a segment. We use a common averaging scheme in the optimization literature \citep{martens2015optimizing,george2018fast,gupta2018shampoo} to compute the averaged EK-FAC factors. We first compute the activation covariance matrices $\mathbf{A}$ and pseudo-gradient covariance matrices $\mathbf{S}$ for all model checkpoints. These matrices are obtained by computing the statistics over all data points once ($1$ epoch). Then, we take the average over these covariance matrices to obtain $\bar{\mathbf{A}} = \frac{1}{C_\ell} \sum_{k=1}^{C_\ell} \mathbf{A}_k$ and $\bar{\mathbf{S}} = \frac{1}{C_\ell} \sum_{k=1}^{C_\ell} \mathbf{S}_k$, where $C_\ell$ is the total number of model checkpoints for the $\ell$-th segment and $\mathbf{A}_k$ and $\mathbf{S}_k$ are covariance matrices for the $k$-th checkpoint. Then, we perform eigendecomposition on these averaged covariance matrices to obtain the eigenvectors $\bar{\mathbf{Q}}_\mathbf{A}$ and $\bar{\mathbf{Q}}_\mathbf{S}$. Under the eigenbasis $\bar{\mathbf{Q}}_\mathbf{A} \otimes \bar{\mathbf{Q}}_\mathbf{S}$, we compute the corrected eigenvalues $\boldsymbol{\Lambda}_k$ for each model checkpoint (\Cref{eq:ekfac_eigen}) and then average the eigenvalues to obtain $\bar{\boldsymbol{\Lambda}}$. In summary, the averaged (Gauss-Newton) Hessian for a particular segment is approximated as:
\begin{align}
    \bar{\mathbf{G}} \approx (\bar{\mathbf{Q}}_\mathbf{A} \otimes \bar{\mathbf{Q}}_\mathbf{S}) \bar{\boldsymbol{\Lambda}} (\bar{\mathbf{Q}}_\mathbf{A} \otimes \bar{\mathbf{Q}}_\mathbf{S})^\top.
\end{align}

\methods requires computing the covariance matrices and corrected eigenvalues for each model checkpoint. Moreover, calculating the TDA scores for all training data points requires computing the training gradients $C$ times, where $C$ is the total number of checkpoints. Hence, \methods is approximately $C$ times more computationally expensive than influence functions evaluated at the final checkpoint. In \Cref{subsec:algo}, we introduced a more efficient variant, which averages the parameters within a segment instead. This variant only needs to compute the EK-FAC factors once for each segment and requires computing the EK-FAC factors and gradients $L$ times. Hence, it is $L$ times more computationally expensive than influence functions.

When the model is trained with SGD with a heavy ball momentum $\beta$ (SGDm), we scaled the learning rate used in \methods as $\barLR_\ell (1 - \beta)^{-1}$ to account for the effective learning rate (terminal velocity). In cases where AdamW optimizers are used as in \Cref{app:general}, computing the matrix exponential for $\bar{\mathbf{P}}_\ell^{1/2} \bar{\hessian}_\ell \bar{\mathbf{P}}_\ell^{1/2}$ is challenging with EK-FAC. We additionally keep track of the diagonal Hessian approximation (which can be easily and efficiently obtained when computing the corrected eigenvalues in \Cref{eq:ekfac_eigen}) and use the diagonal Hessian approximation for computing the matrix exponential in \Cref{eqn:adam} and \Cref{eq:precond_r2}. Note that we still use the EK-FAC factors to compute $\barHess_\ell^{-1} \barGrad_\ell$ in \Cref{eq:precond_r2}.

\section{Additional Results}
\label{app:additional_experiments}

In this section, we present additional experimental results, including a comparison with additional baseline TDA techniques (\Cref{app:add_baseline}), an LDS evaluation of a computationally faster variant of \textsc{Source} (\Cref{app:avg_params}), counterfactual evaluation on linear models (\Cref{app:counterfactual_linear}), and visualizations of top positively and negatively influential training data points for each TDA technique (\Cref{app:qualti}).

\subsection{Additional Baseline Comparisons}
\label{app:add_baseline}

\begin{table}[]
\centering
\begin{tabular}{@{}lcc@{}}
\toprule
\multirow{2}{*}{Methods} & \multicolumn{2}{c}{LDS} \\ \cmidrule(l){2-3} 
 & Single Model & Multiple Models \\ \midrule
\textsc{RepSim} \citep{caruana1999case} & $0.03 \pm 0.02$ & $0.04 \pm 0.02$ \\
\textsc{TracIn} \citep{pruthi2020estimating} & $0.20 \pm 0.02$ & $0.21 \pm 0.03$ \\
\textsc{Trak} \citep{park2023trak} & $0.08 \pm 0.01$ & $0.26 \pm 0.00$ \\
\textsc{IF} \citep{koh2017understanding,grosse2023studying} & $0.30 \pm 0.01$ & $0.45 \pm 0.01$ \\
\textsc{Downsampling} \citep{feldman2020neural} & - & $0.11 \pm 0.02$ \\
\textsc{Hydra} \citep{chen2021hydra} & $0.16 \pm 0.02$ & $0.17 \pm 0.02$ \\
\textsc{Source} with averaged parameters (\textbf{ours}) & $0.42 \pm 0.01$ & $0.48 \pm 0.02$ \\
\textsc{Source} (\textbf{ours}) & $\textbf{0.46} \boldsymbol{\pm} \textbf{0.01}$ & $\textbf{0.53} \boldsymbol{\pm} \textbf{0.01}$ \\ \bottomrule
\end{tabular}
\vspace{-0.2cm}
\caption{Linear datamodeling scores (LDS) at $\alpha = 0.5$ for \methods ($L = 3$) and baseline TDA techniques (including \textsc{Downsampling} and \textsc{Hydra}) on the FashionMNIST dataset. We show the $95\%$ bootstrap confidence intervals.}

\label{tab:more_baseline}
\end{table}

We compare \methods with empirical influence (\textsc{Downsampling}) \citep{feldman2020neural} and the fast version of \textsc{Hydra} \citep{chen2021hydra} on the FashionMNIST task. Results for these techniques on other tasks were omitted, since \textsc{Downsampling} requires retraining the model over $500$ times and \textsc{Hydra} necessitates saving all intermediate checkpoints throughout training. The implementation details are provided in \Cref{app:baseline}, and the results are shown in \Cref{tab:more_baseline}. \methods achieves the highest LDS on both single and multiple model setups compared to existing baseline TDA techniques we considered. 

\subsection{\textsc{Source} with Averaged Parameters}
\label{app:avg_params}

\begin{figure*}[!t]
    \centering
    \includegraphics{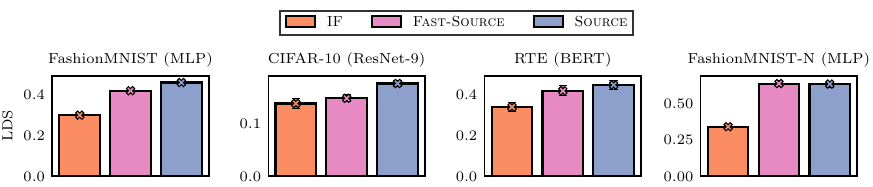}
    \vspace{-0.2cm}
    \caption{Linear datamodeling scores (LDS) at $\alpha = 0.5$ for influence functions, \textsc{Fast-Source} (see \Cref{app:avg_params}), and \method. The LDS is shown for a single model (single-training-run) setup.}
    \label{fig:average_lds}
\end{figure*}

In \Cref{subsec:algo}, we introduced a more computationally efficient version of \textsc{Source}, which averages the parameters within a segment instead of Hessians and gradients. Here, we present the LDS results at $\alpha = 0.5$ for the faster version, termed \textsc{Fast-Source}, for FashionMNIST, CIFAR-10, RTE, and FashionMNIST-N tasks. The results are shown in \Cref{fig:average_lds}. We observe that \textsc{Fast-Source} outperforms influence functions on these tasks, while it generally achieves a lower LDS compared to \textsc{Source}.

\subsection{Counterfactual Evaluations on Linear Models}
\label{app:counterfactual_linear}

\begin{figure*}[t]
    \vspace{-0.4cm}
    \centering
    \includegraphics{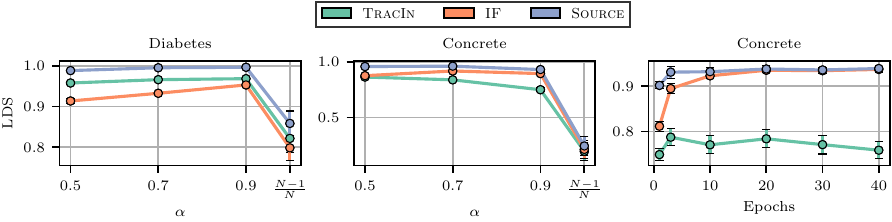}
    \vspace{-0.2cm}
    \caption{(Left \& Middle) Linear datamodeling scores (LDS) for various values of data sampling ratios $\alpha$ on linear regression and logistic regression tasks trained for $3$ epochs. (Right) The LDS at $\alpha = 0.9$ for models trained with varying numbers of epochs. The error bars show $95\%$ bootstrap confidence intervals.}
    \label{fig:linear_lds}
\end{figure*}

\begin{figure*}[t]
    \centering
    \includegraphics{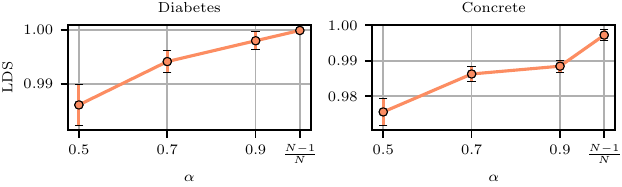}
    \vspace{-0.2cm}
    \caption{Linear datamodeling scores (LDS) on linear regression and logistic regression tasks for influence functions when TDA is performed on the optimal solution.}
    \label{fig:optimal_lds}
\end{figure*}

In this section, we demonstrate the effectiveness of \methods on linear models when the model has not been trained until convergence. We trained linear regression on the Concrete dataset and logistic regression on the Diabetes dataset \citep{smith1988using} for $3$ epochs with a batch size of $32$. We also constructed the LDS ground truth using SGD with the same hyperparameters. We applied \textsc{TracIn}, \textsc{IF}, and \methods (with $L = 1$) to the trained model and computed the LDS for various data sampling ratios $\alpha$. The results are shown in \Cref{fig:linear_lds} (Left \& Middle). \methods achieves higher LDS on all data sampling ratios for both regression and classification tasks. We further show the LDS at $\alpha = 0.9$ with varying numbers of epochs in \Cref{fig:linear_lds} (Right). (The LDS ground truth is recomputed at each epoch.) We observe a larger LDS gap between \textsc{Source} and \textsc{IF} when the model was only trained for a small number of epochs, and the gap reduces as we train the model for a larger number of iterations. These results show that our formulation for \textsc{Source} better supports TDA when the network has not fully converged, even in the case of linear models.

For completeness, we show the LDS for influence functions when the TDA is performed on the optimal solution in \Cref{fig:optimal_lds}. For each model, we computed the optimal solution (for logistic regression, we used the L-BFGS \citep{liu1989limited}), computed the influence function estimates, and evaluated their accuracy with LDS (also obtained by computing the optimal solution without some data points). As shown in \Cref{fig:optimal_lds}, influence functions obtain high correlations with the ground truth across various values of data sampling ratio $\alpha$. In contrast to neural network experiments in \Cref{subsec:lds_eval}, we observe an increase in the LDS as the data sampling ratio $\alpha$ increases (predicting the effect of removing a smaller number of data points), as the group influence predictions introduce more approximation error \citep{bae2022if}. Notably, we obtain a high LDS when $\alpha = \sfrac{(N-1)}{N}$ (removing a single data point), as the LDS is computed at the precise optimal solution (see \Cref{app:loo} for the discussion). \textsc{TracIn} and \textsc{Source} are not applicable in these contexts, as we computed the optimal solution with the direct solution or with L-BFGS, instead of with gradient descent.

\subsection{Qualitative Results}
\label{app:qualti}

\begin{figure*}[!t]
    \vspace{-1cm}
    \centering
    \resizebox{\textwidth}{!}{%
    \begin{tabular}{p{\textwidth}}
        \textsc{RepSim}\\
        \midrule
        \centering
    \includegraphics{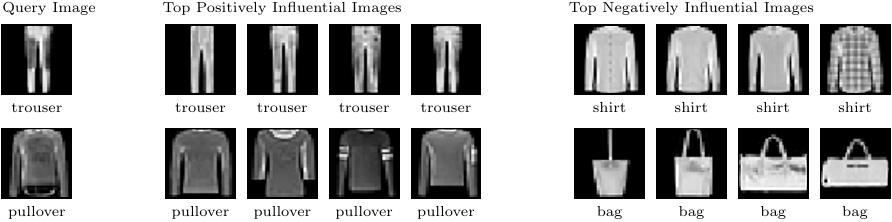}
    \end{tabular}
    }
    \resizebox{\textwidth}{!}{%
    \begin{tabular}{p{\textwidth}}
        \textsc{TracIn}\\
        \midrule
        \centering
    \includegraphics{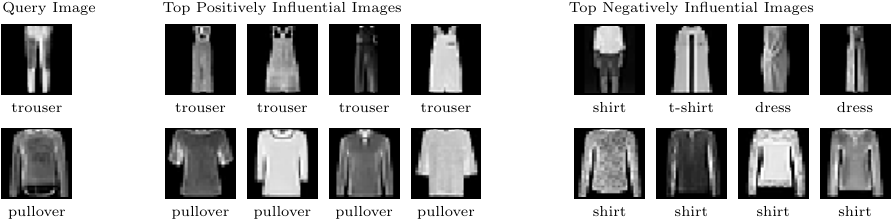}
    \end{tabular}
    }
    \resizebox{\textwidth}{!}{%
    \begin{tabular}{p{\textwidth}}
        \textsc{Trak}\\
        \midrule
        \centering
    \includegraphics{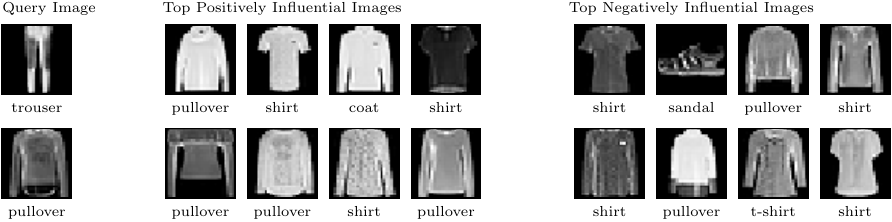}
    \end{tabular}
    }
    \resizebox{\textwidth}{!}{%
    \begin{tabular}{p{\textwidth}}
        \textsc{IF}\\
        \midrule
        \centering
    \includegraphics{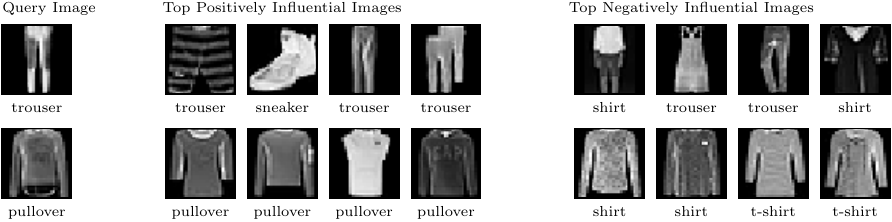}
    \end{tabular}
    }
    \resizebox{\textwidth}{!}{%
    \begin{tabular}{p{\textwidth}}
        \textsc{Source}\\
        \midrule
        \centering
    \includegraphics{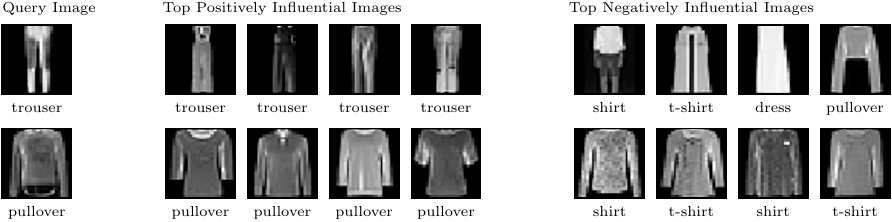}
    \end{tabular}
    }
    \vspace{-0.4cm}
    \caption{Top positively and negatively influential training images identified by \methods and baseline TDA techniques on the FashionMNIST dataset.}
    \label{fig:fmnist}
\end{figure*}

\begin{figure*}[!t]
    \vspace{-1cm}
    \centering
    \resizebox{\textwidth}{!}{%
    \begin{tabular}{p{\textwidth}}
        \textsc{RepSim}\\
        \midrule
        \centering
    \includegraphics{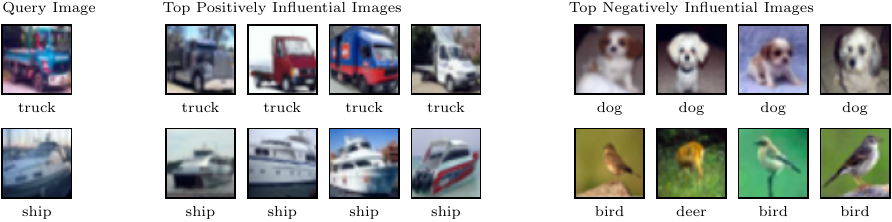}
    \end{tabular}
    }
    \resizebox{\textwidth}{!}{%
    \begin{tabular}{p{\textwidth}}
        \textsc{TracIn}\\
        \midrule
        \centering
    \includegraphics{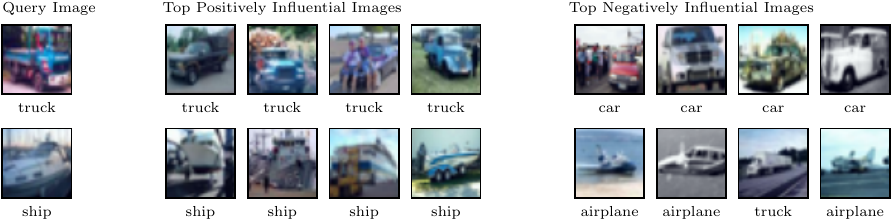}
    \end{tabular}
    }
    \resizebox{\textwidth}{!}{%
    \begin{tabular}{p{\textwidth}}
        \textsc{Trak}\\
        \midrule
        \centering
    \includegraphics{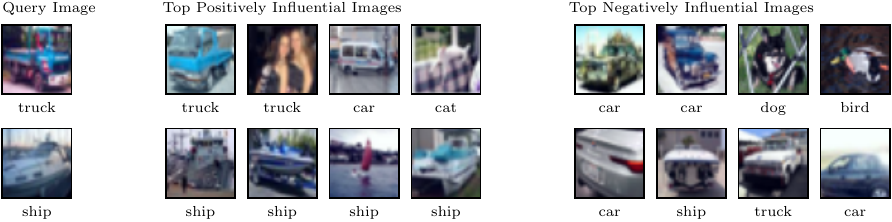}
    \end{tabular}
    }
    \resizebox{\textwidth}{!}{%
    \begin{tabular}{p{\textwidth}}
        \textsc{IF}\\
        \midrule
        \centering
    \includegraphics{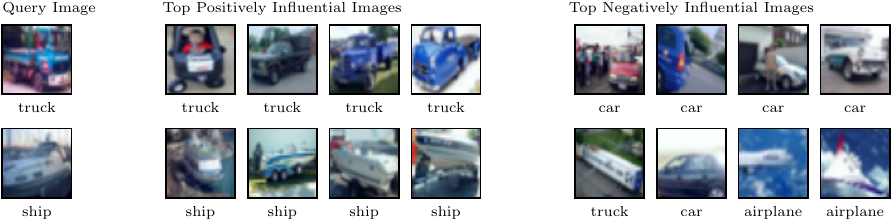}
    \end{tabular}
    }
    \resizebox{\textwidth}{!}{%
    \begin{tabular}{p{\textwidth}}
        \textsc{Source}\\
        \midrule
        \centering
    \includegraphics{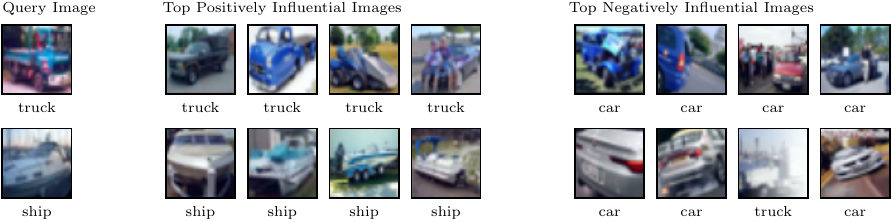}
    \end{tabular}
    }
    \vspace{-0.4cm}
    \caption{Top positively and negatively influential training images identified by \methods and baseline TDA techniques on the CIFAR-10 dataset. Note that we labeled the ``automobile'' class as ``car''.}
    \label{fig:cifar}
\end{figure*}

\begin{figure*}[!t]
    \vspace{-1cm}
    \centering
    \resizebox{\textwidth}{!}{%
    \begin{tabular}{p{\textwidth}}
        \textsc{RepSim}\\
        \midrule
        \centering
    \includegraphics{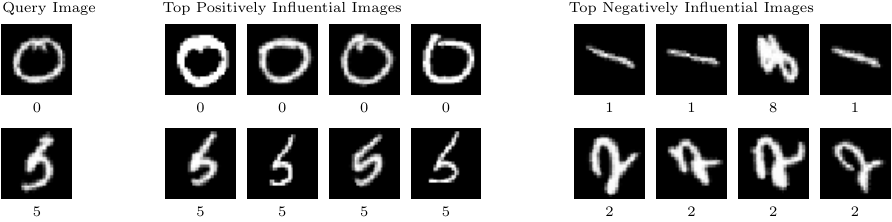}
    \end{tabular}
    }
    \resizebox{\textwidth}{!}{%
    \begin{tabular}{p{\textwidth}}
        \textsc{TracIn}\\
        \midrule
        \centering
    \includegraphics{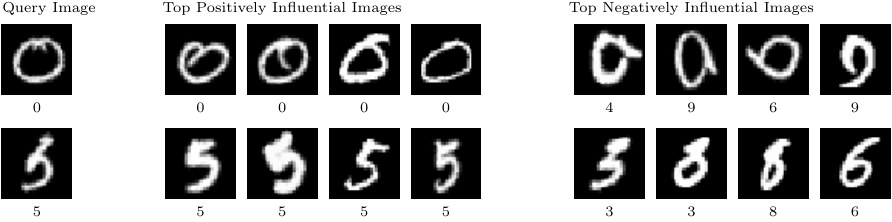}
    \end{tabular}
    }
    \resizebox{\textwidth}{!}{%
    \begin{tabular}{p{\textwidth}}
        \textsc{Trak}\\
        \midrule
        \centering
    \includegraphics{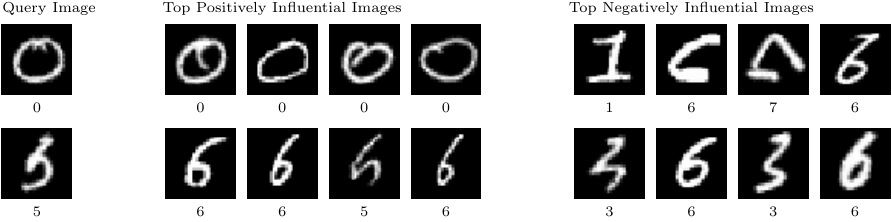}
    \end{tabular}
    }
    \resizebox{\textwidth}{!}{%
    \begin{tabular}{p{\textwidth}}
        \textsc{IF}\\
        \midrule
        \centering
    \includegraphics{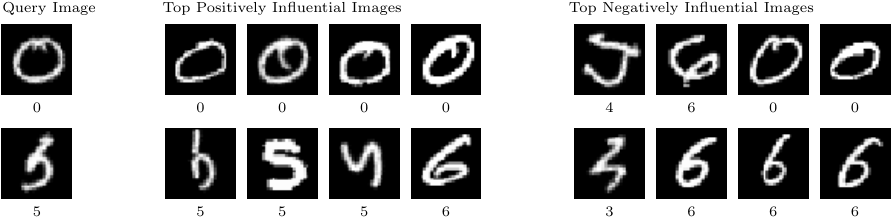}
    \end{tabular}
    }
    \resizebox{\textwidth}{!}{%
    \begin{tabular}{p{\textwidth}}
        \textsc{Source}\\
        \midrule
        \centering
    \includegraphics{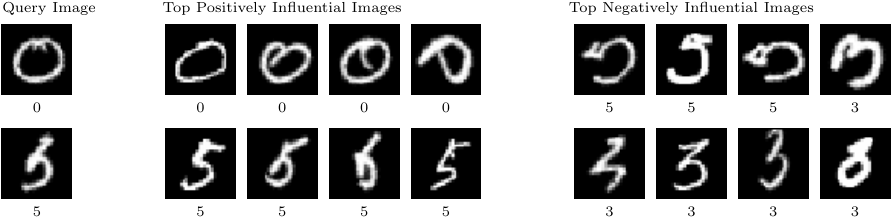}
    \end{tabular}
    }
    \vspace{-0.4cm}
    \caption{Top positively and negatively influential training images identified by \methods and baseline TDA techniques on the RotatedMNIST dataset.}
    \label{fig:rotatedmnist}
\end{figure*}

\begin{figure*}[!t]
    \vspace{-1cm}
    \centering
    \resizebox{\textwidth}{!}{%
    \begin{tabular}{p{\textwidth}}
        \textsc{RepSim}\\
        \midrule
        \centering
    \includegraphics{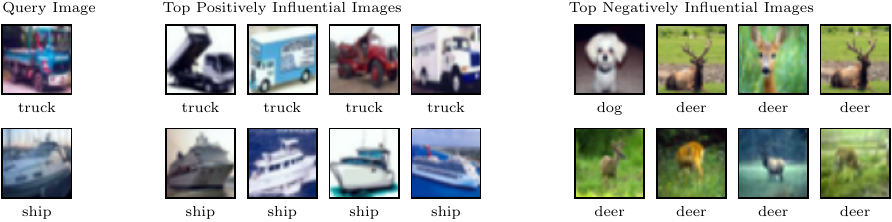}
    \end{tabular}
    }
    \resizebox{\textwidth}{!}{%
    \begin{tabular}{p{\textwidth}}
        \textsc{TracIn}\\
        \midrule
        \centering
    \includegraphics{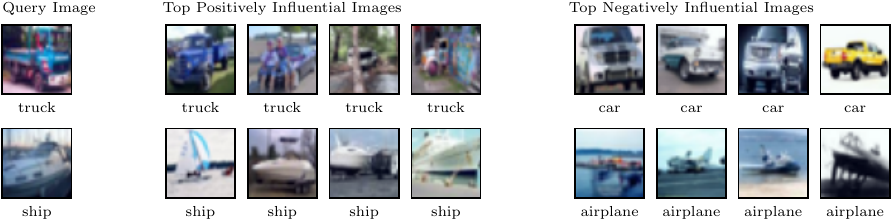}
    \end{tabular}
    }
    \resizebox{\textwidth}{!}{%
    \begin{tabular}{p{\textwidth}}
        \textsc{Trak}\\
        \midrule
        \centering
    \includegraphics{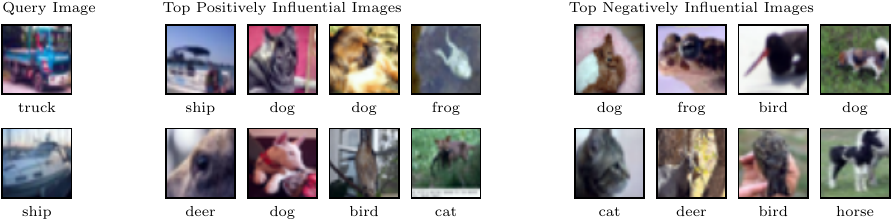}
    \end{tabular}
    }
    \resizebox{\textwidth}{!}{%
    \begin{tabular}{p{\textwidth}}
        \textsc{IF}\\
        \midrule
        \centering
    \includegraphics{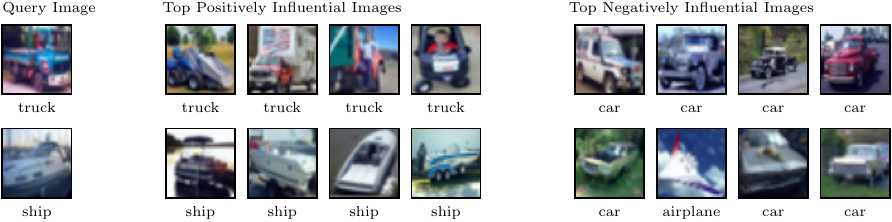}
    \end{tabular}
    }
    \resizebox{\textwidth}{!}{%
    \begin{tabular}{p{\textwidth}}
        \textsc{Source}\\
        \midrule
        \centering
    \includegraphics{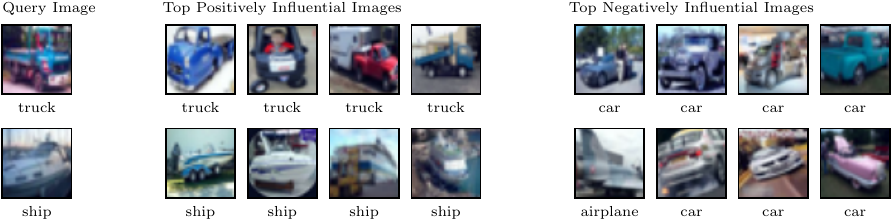}
    \end{tabular}
    }
    \vspace{-0.4cm}
    \caption{Top positively and negatively influential training images identified by \methods and baseline TDA techniques (single model setting) on the CIFAR-10 dataset.}
    \label{fig:cifar_single}
\end{figure*}

We first present the top positively and negatively influential data points obtained by each TDA technique on multiple model settings. Note that for these multiple model settings, \textsc{RepSim}, \textsc{TracIn}, \textsc{Trak}, \textsc{IF}, and \methods use an ensemble of 10 models trained with different random choices. The results for FashionMNIST, CIFAR-10, and RotatedMNIST are shown in \Cref{fig:fmnist}, \Cref{fig:cifar}, and \Cref{fig:rotatedmnist}, respectively. We also show the top positively and negatively influential data points on the CIFAR-10 dataset for a single model setup in \Cref{fig:cifar_single}. In \Cref{fig:rte}, we present the top positively and negatively influential data points obtained by \methods on the RTE dataset.

\begin{table}[p]
    \tiny
    \resizebox{\columnwidth}{!}{%
    \begin{tabular}{  p{3.4cm}  p{3.4cm}  p{3.4cm} }
        \toprule
\textbf{Query Data Point}      
& \textbf{Top Positively Influential Data Point}   
& \textbf{Top Negatively Influential Data Point} \\\midrule
Dana Reeve, the widow of the actor Christopher Reeve, has died of lung cancer at age 44, according to the Christopher Reeve Foundation. / Christopher Reeve had an accident. \textbf{(not entailment)}
& Though fearful of a forthcoming performance evaluation by her boss, Zoe must unravel the life of a man just found dead of a heart attack, who was supposed to have died three years earlier in a boating accident. / Zoe died in a boating accident. \textbf{(not entailment)}     
& Actor Christopher Reeve, best known for his role as Superman, is paralyzed and cannot breathe without the help of a respirator after breaking his neck in a riding accident in Culpeper, Va., on Saturday. / Christopher Reeve had an accident. \textbf{(entailment)} \\ \\ \hline
\\ Yet, we now are discovering that antibiotics are losing their effectiveness against illness. Disease-causing bacteria are mutating faster than we can come up with new antibiotics to fight the new variations. / Bacteria is winning the war against antibiotics. \textbf{(entailment)}
& The papers presented show that all European countries are experiencing rapidly aging populations that will cause sharp increases in the cost of retirement income over the next several decades. / National pension systems currently adopted in Europe are in difficulties. \textbf{(entailment)}    
& Humans have won notable battles in the war against infection - and antibiotics are still powerful weapons - but nature has evolution on its side, and the war against bacterial diseases is by no means over. / Bacteria is winning the war against antibiotics. \textbf{(not entailment)} \\ \\\hline
\\
Security forces were on high alert after an election campaign in which more than 1,000 people, including seven election candidates, have been killed. / Security forces were on high alert after a campaign marred by violence. \textbf{(entailment)}
& Police sources stated that during the bomb attack involving the Shining Path, two people were injured. / Two people were wounded by a bomb. \textbf{(entailment)}     
& Pakistan President Pervez Musharraf has ordered security forces to take firm action against rioters following the assassination of opposition leader Benazir Bhutto. The violence has left at least 44 people dead and dozens injured. Mr. Musharraf insisted the measures were to protect people. VOA's Ayaz Gul reports from Islamabad that a bitter dispute has also erupted over how the 54-year-old politician died and who was behind her assassination. / Musharraf has ordered rioters to take firm action against security forces. \textbf{(not entailment)} \\ \\ \hline
\\
In 1979, the leaders signed the Egypt-Israel peace treaty on the White House lawn. Both President Begin and Sadat received the Nobel Peace Prize for their work. The two nations have enjoyed peaceful relations to this day. / The Israel-Egypt Peace Agreement was signed in 1979. \textbf{(entailment)}
& Following the Israel-Egypt Peace Treaty of 1979, Israel agreed to withdraw from the Sinai Peninsula, in exchange for peace with its neighbor. For over two decades, the Sinai Peninsula was home to about 7,000 Israelis. / The Israel-Egypt Peace Agreement was signed in 1979. \textbf{(entailment)}      
& Canada and the United States signed an agreement on January 30, 1979, to amend the treaty to allow subsistence hunting of waterfowl. / The Israel-Egypt Peace Agreement was signed in 1979. \textbf{(not entailment)} \\ \\
        \bottomrule
    \end{tabular}
    }
        \vspace{-0.2cm}
    \caption{Top positively and negatively influential data points identified by \methods on the RTE dataset. A data point in the RTE dataset consists of a pair of sentences (separated by a forward slash ``/'') and a label indicating whether the second sentence entails the first sentence (entailment) or not (not entailment).}
    \label{fig:rte}
\end{table}

\section{Limitations of \textsc{Source}}
\label{app:linmitations}

Compared to the influence function employing the same EK-FAC parameterization \citep{grosse2023studying}, the practical implementation of the \methods requires the computation of EK-FAC factors and gradients for all checkpoints (when performing TDA on all segments). Denoting the total number of checkpoints as $C$ and the total number of segments as $L$, \methods exhibits an approximate computational cost of $C$ times higher. Our experiments used configurations with $C = 6$ and $L = \{2, 3\}$. We also introduced a faster version of \methods in \Cref{app:avg_params}, which directly averages the parameters instead of averaging the EK-FAC factors and gradients; the faster version is $L$ times computationally expensive compared to the EK-FAC influence functions.

Compared to implicit-differentiation-based TDA techniques, \methods requires access to intermediate checkpoints throughout the training process and corresponding hyperparameters such as learning rate, number of iterations, and preconditioning matrix. In cases where the details of the training process are not available, implicit-differentiation-based TDA techniques, such as \textsc{Trak} \citep{park2023trak} and influence functions \citep{koh2017understanding,grosse2023studying}, may be preferable.

Moreover, \methods approximates the distributions of the Hessian and gradient as stationary within each segment of the training trajectory. In certain scenarios, this may not be a reasonable approximation. For instance, when pre-training large transformer models, the Hessian or gradients may undergo drastic changes throughout the training process. If the stationarity approximation is too inaccurate, one can enhance the fidelity of \methods by dividing the training trajectory into a larger number of segments, albeit at the cost of increased computational requirements. While we used a fixed number of segments and checkpoints, partitioned equally at the early, middle, and late stages of training, we can extend \methods by automatically determining when to segment by examining the changes in the Hessian or gradients, which we leave for future work.

\end{appendices}

\end{document}